\documentclass{article}

\usepackage[utf8]{inputenc}
\usepackage[english]{babel}

\usepackage{caption}

\usepackage[letterpaper,top=2cm,bottom=2cm,left=3cm,right=3cm,marginparwidth=1.75cm]{geometry}

\usepackage{booktabs}
\usepackage{multirow}
\usepackage[table,xcdraw]{xcolor}
\usepackage{makecell} 

\usepackage{amsmath}
\usepackage{graphicx}
\usepackage{booktabs}
\usepackage{multicol}
\usepackage{multirow}
\usepackage[colorlinks=true, allcolors=blue]{hyperref}
\usepackage{xurl}
\usepackage{amsmath}
\usepackage{amssymb}
\usepackage{amsfonts}

\usepackage{listings}
\usepackage{xcolor}
\usepackage{longtable}

\definecolor{codegreen}{rgb}{0,0.6,0}
\definecolor{codegray}{rgb}{0.5,0.5,0.5}
\definecolor{codepurple}{rgb}{0.58,0,0.82}
\definecolor{backcolour}{rgb}{0.95,0.95,0.92}

\lstdefinestyle{mystyle}{
    backgroundcolor=\color{backcolour},   
    commentstyle=\color{codegreen},
    keywordstyle=\color{magenta},
    numberstyle=\tiny\color{codegray},
    stringstyle=\color{codepurple},
    basicstyle=\ttfamily\footnotesize,
    breakatwhitespace=false,         
    breaklines=true,                 
    captionpos=b,                    
    keepspaces=true,                 
    numbers=left,                    
    numbersep=5pt,                  
    showspaces=false,                
    showstringspaces=false,
    showtabs=false,                  
    tabsize=2
}
\usepackage{listings}
\usepackage{xcolor}
\usepackage[T1]{fontenc}

\definecolor{deepblue}{rgb}{0,0,0.5}
\definecolor{deepred}{rgb}{0.6,0,0}
\definecolor{deepgreen}{rgb}{0,0.5,0}

\lstnewenvironment{python}[1][] 
{\lstset{language=Python, #1}} 
{}

\title{MolClaw: An Autonomous Agent with Hierarchical Skills for Drug Molecule Evaluation, Screening, and Optimization}
\author{Lisheng Zhang$^{1\ast}$, Lilong Wang$^{2\ast}$, Xiangyu Sun$^{2\ast}$, Wei Tang$^{3\ast}$, Haoyang Su$^{2\ast}$, Yuehui Qian$^{1}$, \\Qikui Yang$^{1}$, Qingsong Li$^{1}$,  Zhenyu Tang$^{2}$, Haoran Sun$^{2}$, Yingnan Han$^{2}$, Yankai Jiang$^{2}$, \\Wenjie Lou$^{2}$, Bowen Zhou$^{2}$, Xiaosong Wang$^{2\dagger}$, Lei Bai$^{2\dagger}$, Zhengwei Xie$^{1\dagger}$}

\begin{document}
\maketitle

\vspace{-3em}
\begin{center}
    \begin{tabular}{c}
        $^{1}$Peking University Health Science Center, Peking University, Beijing, China \\
        $^{2}$Shanghai AI Laboratory, Shanghai, China \\
        $^{3}$Academy for Advanced Interdisciplinary Studies, Peking University, Beijing, China \\
        $\ast$ These authors contributed equally \\
        $\dagger$ Corresponding author \\
        Corresponding author(s): wangxiaosong@pjlab.org.cn, bailei@pjlab.org.cn, xiezhengwei@hsc.pku.edu.cn\\
    \end{tabular}
\end{center}

\begin{abstract}

\noindent Computational drug discovery, particularly the complex workflows of drug molecule screening and optimization, requires orchestrating dozens of specialized tools in multi-step workflows, yet current AI agents struggle to maintain robust performance and consistently underperform in these high-complexity scenarios. 
Here, we present \textbf{MolClaw}, an autonomous agent that orchestrates over 30 resources, including specialized databases, computational tools, and predictive models, through a hierarchical skill ecosystem. Methodology-level principles spanning eight domains provide overarching scientific guidance, 14 workflow-level skills assemble validated pipelines with quality control and reflective feedback, and 60 tool-level skills standardize atomic operations, complemented by optional deep research capabilities for literature retrieval and verification. Together, these components enable MolClaw to execute tasks in structured and reproducible procedures, perform multi-round quality checks, and iteratively refine strategies. To benchmark its performance, we introduce \textbf{MolBench}, covering molecular screening, optimization, and end-to-end discovery challenges involving more than 50 sequential tool calls. MolClaw achieves state-of-the-art results across all metrics, with ablation studies indicating that gains are concentrated on tasks requiring structured workflows, while tasks solvable by ad hoc scripting show minimal improvement. These findings underscore workflow orchestration as a critical bottleneck in AI-driven drug discovery and demonstrate MolClaw’s capability for complex multi-target molecular optimization. Furthermore, MolClaw shows potential for autonomous scientific discovery, execution, and skill generation.
\end{abstract}

\section*{Introduction}
\label{sec:introduction}
The discovery of new therapeutic molecules remains one of the most resource-intensive endeavors in modern science. Bringing a single drug from initial concept to regulatory approval typically requires 10 to 15 years of effort and over US\$1 to 2 billion in investment, yet fewer than one in ten candidates entering Phase I clinical trials ultimately reach the market \cite{sun202290,wouters2020estimated}. Central to this attrition is the early-stage computational drug discovery pipeline, a complex, multi-step workflow spanning target structure determination, binding pocket identification, virtual screening, \textit{de novo} molecular generation, molecular docking, binding free energy estimation, ADMET profiling, and iterative lead optimization \cite{sadybekov2023computational,vamathevan2019applications}. Each step relies on specialized software tools with disparate input/output formats, heterogeneous parameter spaces, and steep learning curves: a typical structure-based virtual screening campaign may require coordinated use of ESMFold or AlphaFold for protein structure prediction \cite{lin2023evolutionary,jumper2021highly}, fpocket or P2Rank for pocket detection \cite{le2009fpocket, krivak2018p2rank}, AutoDock Vina or Glide for molecular docking \cite{eberhardt2021autodock,friesner2004glide}, RDKit for physicochemical property calculation \cite{rdkit}, and GROMACS for molecular dynamics refinement \cite{abraham2015gromacs}, often demanding days of manual effort to orchestrate the format conversions, parameter tuning, and quality control between each stage \cite{dharmasivam2025leading}. This fragmentation not only slows the pace of discovery but also introduces reproducibility challenges, as subtle differences in preprocessing steps or parameter choices can substantially alter downstream results \cite{bender2021artificial}.

The rapid progress of large language models (LLMs) over the past two years has opened new avenues for automating complex scientific workflows. Foundation models such as GPT~\cite{achiam2023gpt}, Claude~\cite{anthropic_claude}, Gemini~\cite{team2024gemini}, and open-source alternatives including LLaMA \cite{touvron2023llama} and Qwen \cite{bai2023qwen} have demonstrated strong capabilities in natural language process. Intern-S1 \cite{bai2025intern}, a scientific multimodal foundation model, further exhibits high performance in scientific reasoning and cross-modal task processing. When augmented with tool invocation abilities, referred to as tool-augmented LLMs or AI agents \cite{schick2023toolformer, mialon2023augmented}, these models can execute multi-step computational tasks beyond text generation. Architectures such as ReAct \cite{yao2022react}, which interleave reasoning with action execution, enable agents to decompose complex objectives into sequences of tool calls, interpret intermediate results, and adaptively revise strategies. This paradigm is increasingly recognized as transformative for scientific discovery \cite{wang2023scientific, boiko2023autonomous}, positioning AI agents as collaborative partners that integrate computational models, biomedical tools, and experimental platforms \cite{gao2024empowering}. General-purpose scientific agent frameworks, including InternAgent-1.5 \cite{feng2026internagent}, demonstrate sustained multi-step reasoning by coordinating generation, verification, and evolution subsystems across computational and empirical domains. Open-source platforms such as OpenClaw \cite{openclaw} and Claude Code \cite{anthropic_claude_code} further lower the barrier to constructing tool-augmented agents, offering robust environments for tool orchestration, error recovery, and multi-turn reasoning.

Several pioneering systems have applied LLM agents to chemistry and drug discovery. ChemCrow \cite{m2024augmenting} integrated 18 chemistry tools with GPT-4 for tasks including synthesis planning, property prediction, and materials design, demonstrating autonomous insect repellent synthesis. However, it was limited to single-step tool calls and lacked multi-stage workflows essential for computational drug discovery, such as pocket identification, molecular generation, docking, ADMET filtering, and binding energy estimation. Biomni \cite{huang2025biomni} mined tools and protocols from 2,500 publications to construct a broad biomedical action space, generalizing across tasks like gene prioritization and drug repurposing. Its code-based execution, however, relied on ad hoc Python scripts and lacked structured, reusable workflow abstractions required for complex multi-tool pipelines. DrugAgent \cite{liu2024drugagent} automated ML programming for drug–target interaction prediction but did not support molecular-level tasks such as 3D structure manipulation or docking. TxAgent \cite{gao2025txagent}, fine-tuned on 211 biomedical tools, excelled at therapeutic reasoning and personalized treatment recommendations, yet remained limited to clinical pharmacology. ChatInvent \cite{he2026democratising} demonstrated real-world utility in molecular design and synthesis planning but is proprietary and closed-source. Conceptual frameworks such as "Prompt-to-Pill" \cite{vichentijevikj2026prompt} outline end-to-end multi-agent pipelines from target identification to virtual patient recruitment but lack public implementation or benchmark evaluation. Other recent works, including FROGENT \cite{pan2025frogent}, Mozi \cite{cao2026mozi}, and LIDDiA \cite{averly2025liddia}, explore multi-agent coordination and long-horizon task reliability, yet none integrates the full suite of professional computational chemistry tools or provides a systematic, multi-dimensional benchmark for structure-based drug discovery.

\begin{figure}[p]
    \centering\includegraphics[width=0.9\linewidth]{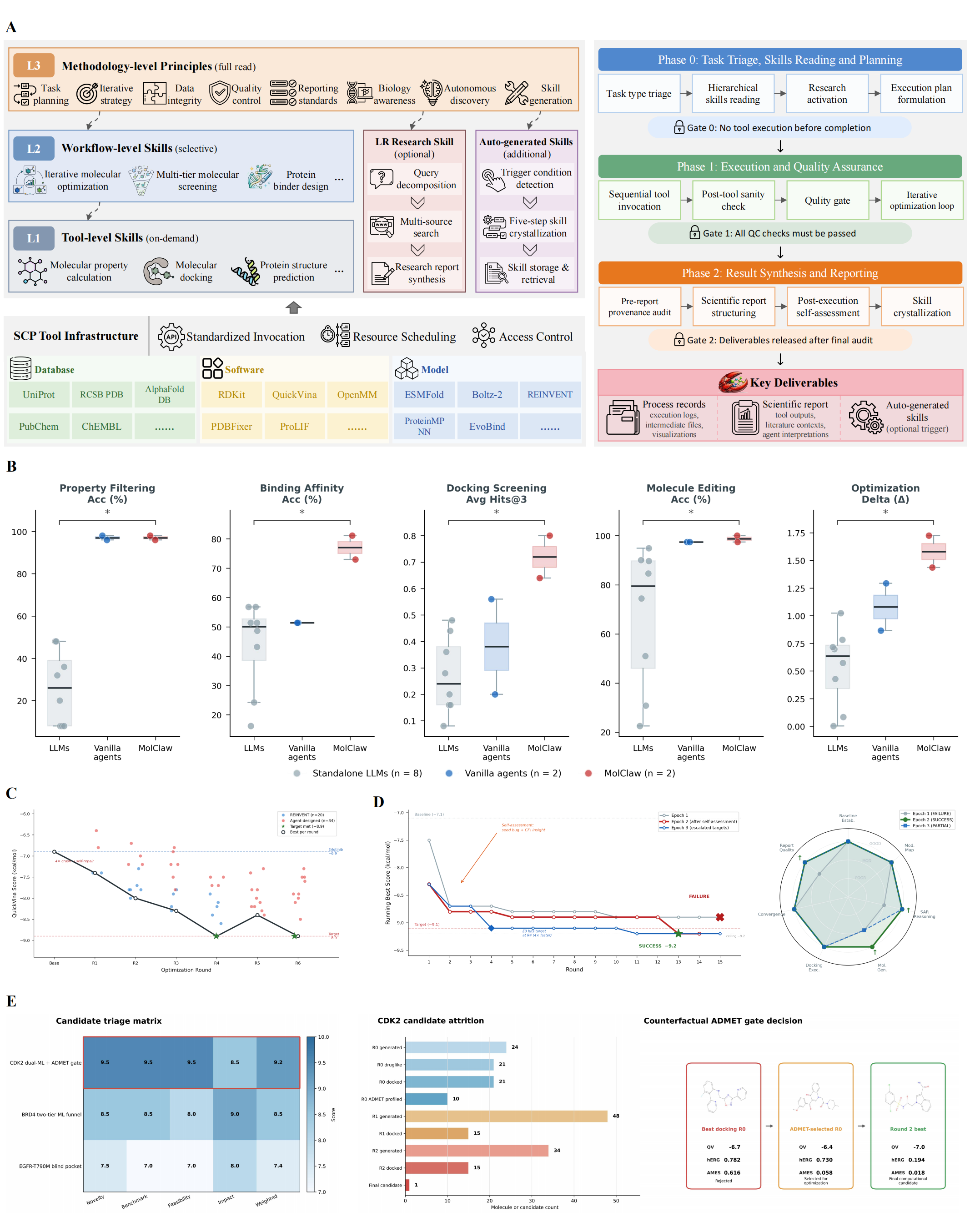}
    \caption{Technical architecture and key capabilities of MolClaw. (\textbf{A}) illustrates the hierarchical skill ecosystem and the MolClaw’s execution process. (\textbf{B}) highlights MolClaw’s superior performance across five molecular screening and optimization tasks. (\textbf{C}, \textbf{D}) demonstrate MolClaw’s ability to conduct long-term, iterative drug discovery processes. (\textbf{E}) showcases autonomous scientific discovery and experimental execution.}
    \label{fig:agent}
\end{figure}

Here, we present MolClaw (shown in Figure~\ref{fig:agent}), an autonomous AI agent for end-to-end early-stage drug discovery, built around a hierarchical skill architecture that encodes computational chemistry expertise into reusable, model-agnostic templates.

MolClaw integrates over 30 professional tools spanning the full early-stage discovery pipeline, from target preparation and binding site identification to molecular generation, docking, dynamics simulation, and ADMET profiling, through a unified SCP infrastructure that handles GPU scheduling, concurrent task management, and standardized access control \cite{jiang2025scp}. Domain expertise is organized into a four-layer skill ecosystem: \textbf{Tool-level (L1)} templates standardize individual tool invocations; \textbf{Workflow-level (L2)} frameworks compose these into end-to-end pipelines for core tasks such as virtual screening and lead optimization, and additionally include meta-workflows for autonomous draft workflow authoring and skill crystallization; \textbf{Methodology-level (L3)} encodes 26 formal scientific principles governing agent decision-making, iterative optimization, and reporting, loaded in full before any execution begins; and a \textbf{Research-level (LR)} layer provides evidence-grounded literature retrieval under a strict computation-first hierarchy. Crucially, an auto-generated skill repository accumulates knowledge crystallized from novel execution traces, allowing MolClaw's capability to expand monotonically with experience without manual curation.

Execution is coordinated through a gate-regulated three-phase framework: mandatory task triage and hierarchical skill loading before any tool call (Phase 0), continuous sanity and cross-validation gating during execution (Phase 1), and post-execution result synthesis coupled with a skill crystallization procedure that converts novel traces into persistent reusable skills (Phase 2). This architecture makes reproducibility, methodological alignment, and self-improvement structural properties of the system rather than emergent behaviors of the underlying model. MolClaw is model- and framework-agnostic, validated with Claude Sonnet 4.6 and Qwen 3.5 across both OpenClaw \cite{openclaw} and Claude Code \cite{anthropic_claude_code} runtimes.

To systematically evaluate MolClaw and establish a community resource for benchmarking drug discovery agents, we introduce MolBench, a multi-dimensional evaluation suite comprising three complementary tiers. MolBench-MS (Molecular Screening) evaluates the agent's ability to correctly filter and rank molecules based on physicochemical constraints and activity predictions. MolBench-MO (Molecular Optimization) assesses the agent's capacity to improve molecular properties through structural modifications while maintaining core scaffolds. MolBench-E2E (End-to-End Discovery) presents three comprehensive drug discovery challenges requiring the agent to autonomously execute extended, multi-phase workflows spanning 8--50+ sequential tool invocations, covering coarse-grained conformational sampling with dual force fields and all-atom reconstruction, multi-round closed-loop molecular property optimization, and structure-guided iterative lead optimization with docking-score-driven convergence. These scenarios closely mirror the complexity and iterative, feedback-driven nature of real-world computational drug discovery campaigns.

We evaluate MolClaw against a comprehensive set of baselines, including frontier LLMs, the domain-specific biomedical agent Biomni \cite{huang2025biomni}, and vanilla agent frameworks (OpenClaw \cite{openclaw} and Claude Code \cite{anthropic_claude_code} without MolClaw skills). Across all three dimensions of MolBench, MolClaw achieves state-of-the-art performance, and ablation studies confirm that the performance gains derive primarily from the hierarchical skill architecture rather than the choice of underlying LLM.

In summary, this work makes the following contributions:

1) \textbf{A comprehensive tool ecosystem for computational drug discovery}. We integrate over 30 professional resources spanning the complete early-stage drug discovery pipeline through a unified SCP infrastructure, enabling fully automated execution of complex multi-tool workflows on GPU-accelerated clusters.

2) \textbf{A hierarchical skill architecture with progressive self-improvement}. We propose a multi-layer skill ecosystem (Tool, Workflow, Methodology, and Research) augmented by a self-expanding auto-generated skill repository, and couple it with a gate-regulated execution framework that enforces methodological rigor at the infrastructure level and enables skill crystallization from novel execution experience. Together, these address the fundamental limitations of flat tool-calling and static knowledge in existing scientific agents.

3) \textbf{MolBench: the first multi-dimensional benchmark for drug discovery agents}. We construct and release MolBench, comprising molecular screening, molecular optimization, and three end-to-end drug discovery challenges that require genuine multi-step, multi-round tool execution across heterogeneous simulation and optimization pipelines, filling a critical gap in the evaluation of agentic systems for drug discovery.

4) \textbf{State-of-the-art performance with systematic evaluation}. MolClaw substantially outperforms all baselines across MolBench dimensions. Ablation studies establish that the hierarchical skill design, not the underlying LLM, is the primary driver of performance, validating the model-agnostic design philosophy and confirming capability transfer across agent runtimes.

\section*{Results}

%
 
\begin{figure}[p]
    \centering\includegraphics[width=\linewidth]{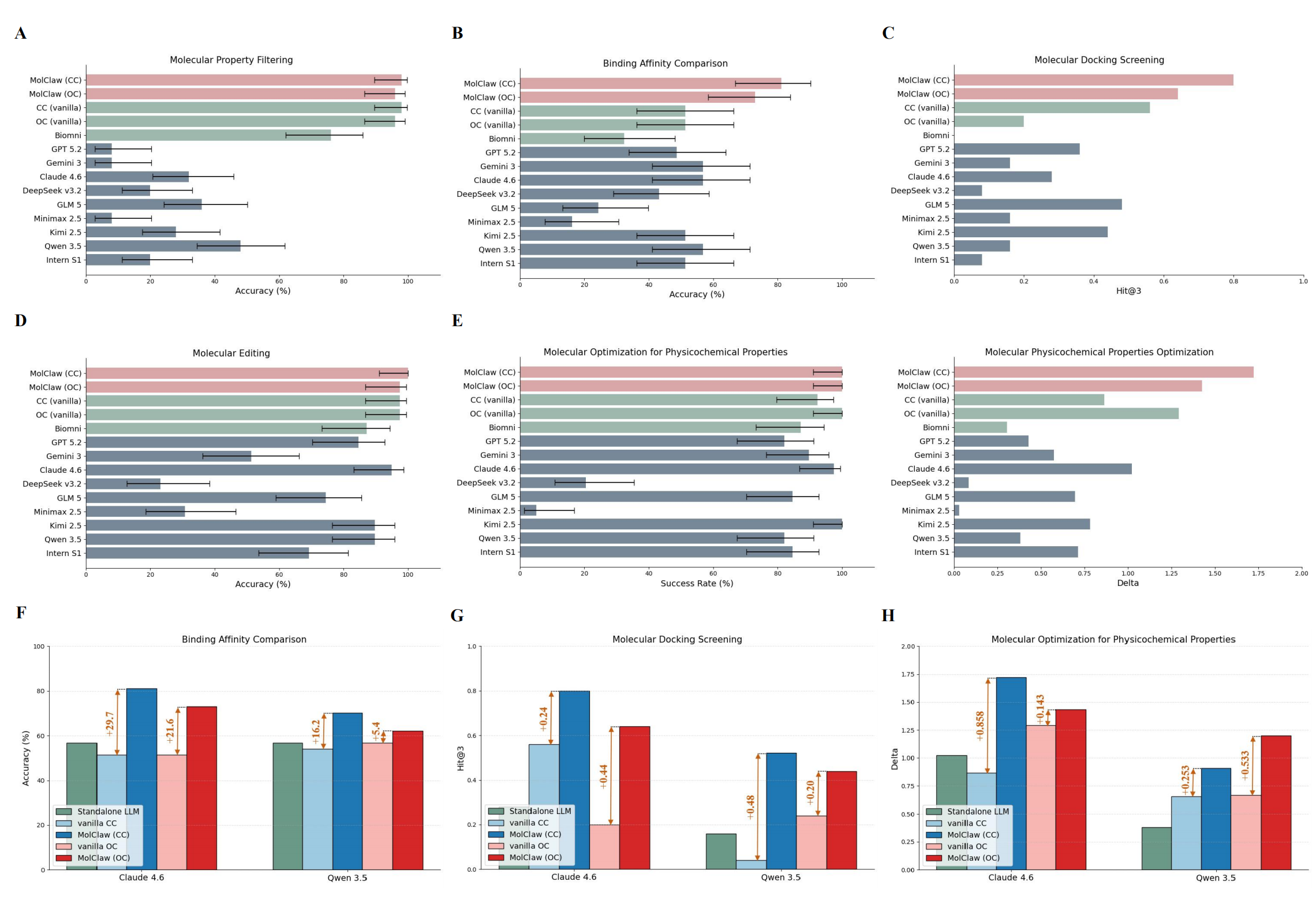}
    \caption{Evaluation results of MolClaw, three representative agents, and multiple large language models (LLMs) on the MolBench-MS and MolBench-MO benchmarks. (\textbf{A--C}) Performance comparison on the three subtasks of MolBench-MS. (\textbf{D}, \textbf{E}) Performance comparison on the two subtasks of MolBench-MO, where both success rate and delta metrics are reported for the second molecular optimization task. (\textbf{F--G}) Ablation study of MolClaw compared with vanilla agents, including Claude Code (CC) and OpenClaw (OC), using the default LLM of Claude Sonnet 4.6 and an alternative open-source LLM of Qwen 3.5.}
    \label{fig:molbench}
\end{figure}

\noindent \textbf{MolClaw achieves state-of-the-art performance across all MolBench-MS and MolBench-MO benchmarks.}
We evaluated MolClaw against nine leading LLMs (GPT~5.2, Claude Sonnet~4.6, Gemini~3, DeepSeek~v3.2, Kimi~2.5, GLM~5, Minimax~2.5, Qwen~3.5, Intern~S1), the biomedical agent Biomni, and two vanilla agent frameworks (Claude Code and OpenClaw without MolClaw skills). Across seven metrics spanning MolBench-MS and MolBench-MO, MolClaw on Claude Code (MolClaw-CC) ranked first or tied for first on every metric, achieving the highest performance on four metrics (binding affinity accuracy, docking hit count, molecule editing accuracy, and optimization delta) and tied best performance on two metrics (property filtering accuracy and optimization success rate) (Fig.~\ref{fig:molbench}A--E). 

MolBench-MS illustrates the value of integrating professional tools. On molecular property filtering ($n = 50$), which requires computing physicochemical descriptors and applying threshold-based constraints, standalone LLMs performed poorly with a mean accuracy of 25.3\% and a range of 8.0 to 48.0\%, confirming their inability to reliably compute molecular descriptors through reasoning alone. All agent-based systems, including vanilla frameworks, achieved near-perfect accuracy of at least 96.0\% by autonomously executing RDKit-based scripts. MolClaw-CC achieved 98.0\%, matching vanilla Claude Code. This control task indicates that tool access, rather than pre-defined skills, limits performance when scripting is sufficient. In contrast, binding affinity comparison ($n = 37$), the most discriminating MolBench task, revealed the limitations of unstructured code generation. Standalone LLMs averaged 45.1\% accuracy, and vanilla agents achieved 51.4\%, while MolClaw-CC reached 81.1\%, a 24.3-point improvement over the best non-MolClaw baseline (Fisher's exact test, $P = 0.043$, Cohen's $h = 0.54$) and a 29.7-point gain over its own vanilla backbone ($P = 0.013$, $h = 0.64$) (Fig.~\ref{fig:molbench}F). This improvement reflects MolClaw’s structured workflow, which coordinates target preparation, binding execution, and scoring into a validated multi-step pipeline incorporating domain expertise that ad hoc approaches cannot replicate. Molecular docking screening ($n = 25$) confirmed this trend, with MolClaw-CC achieving the highest average hit count of 0.80, exceeding vanilla Claude Code at 0.56 and OpenClaw at 0.20, while Biomni failed to produce valid outputs. 

MolBench-MO evaluates molecular design capabilities. Molecule editing ($n = 39$) assesses the ability to perform functional group replacement and scaffold modification. Standalone LLMs showed high variability with mean accuracy of 67.2\%, while both vanilla frameworks reached 97.4\%, and MolClaw-CC achieved perfect accuracy of 100.0\%. Physicochemical property optimization, the most challenging MolBench task, requires iterative improvement of QED, LogP, and LogS. Standalone LLMs averaged $\Delta = 0.521$ with success rates as low as 5.1\%, vanilla frameworks reached $\Delta = 0.866$ to 1.293, and MolClaw-CC achieved the highest $\Delta$ of 1.724 with 100.0\% success. This demonstrates the effectiveness of MolClaw’s L2 iterative workflow, which integrates LLM-guided structural design, property computation, Lipinski violation checking, and automated evaluation to explore chemical space efficiently and reliably.

Furthermore, we performed an ablation study on the three tasks where MolClaw showed relatively modest gains over vanilla agents (Fig.~\ref{fig:molbench}F--H). In this study, we replaced the underlying proprietary LLM with the open-source model Qwen~3.5 for both MolClaw and the vanilla frameworks. Under this replacement, MolClaw consistently outperformed the corresponding vanilla agents across all tasks. These results indicate that the performance improvements are primarily due to MolClaw’s structured skill orchestration and validated workflows, rather than the choice of a particular LLM backbone.
\\[6pt]
\noindent \textbf{MolClaw effectively addresses complex, long-horizon drug discovery tasks.}

 
\begin{figure}[p]
    \centering
    \includegraphics[width=0.8\linewidth]{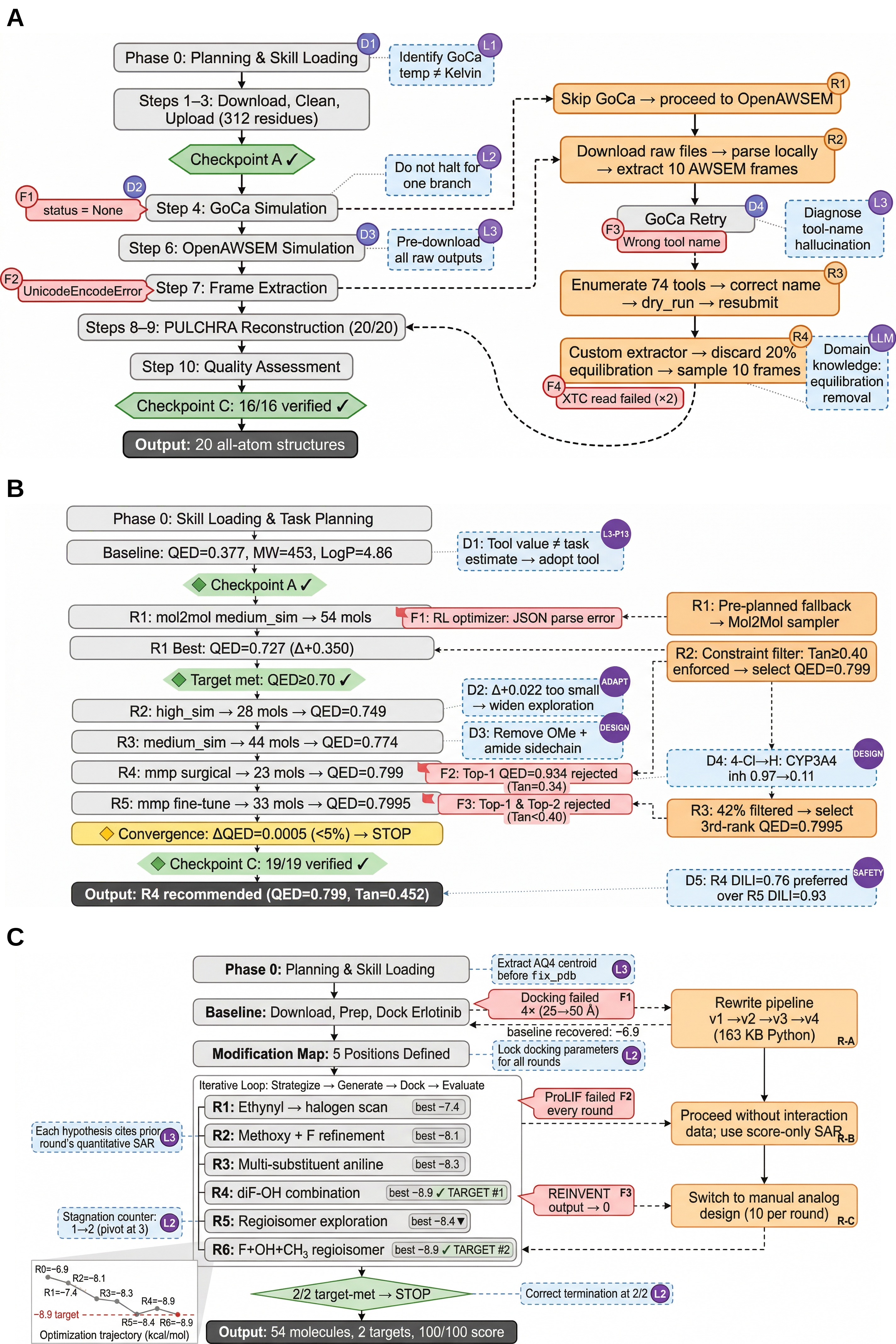}
    \captionsetup{font=footnotesize}
    \caption{\textbf{Agent execution traces for the three MolBench-E2E tasks.}
    (\textbf{A})~E2E-Q1: coarse-grained conformational sampling. Five tool-level failures (red) were resolved via skill-governed recovery actions (orange), yielding 20 verified all-atom structures.
    (\textbf{B})~E2E-Q2: QED-driven iterative optimization. One tool fallback (F1), two constraint-driven rejections (F2--F3), and five strategy adaptations (D1--D5) were autonomously managed across five rounds, with all 19 reported values verified against source files.
    (\textbf{C})~E2E-Q3: structure-guided lead optimization of Erlotinib. Three failure categories---baseline docking crashes, persistent ProLIF errors, and generative-model collapse---were resolved via pipeline rewriting, score-only SAR, and manual analog design. Inset: optimization trajectory reaching the $-8.9$~kcal/mol target.
    Blue dashed boxes: agent decisions; purple badges: governing skill layer; green diamonds: verification checkpoints.}
    \label{fig:e2e-traces}
\end{figure}

\begin{figure}[p]
    \centering
    \includegraphics[width=0.9\linewidth]{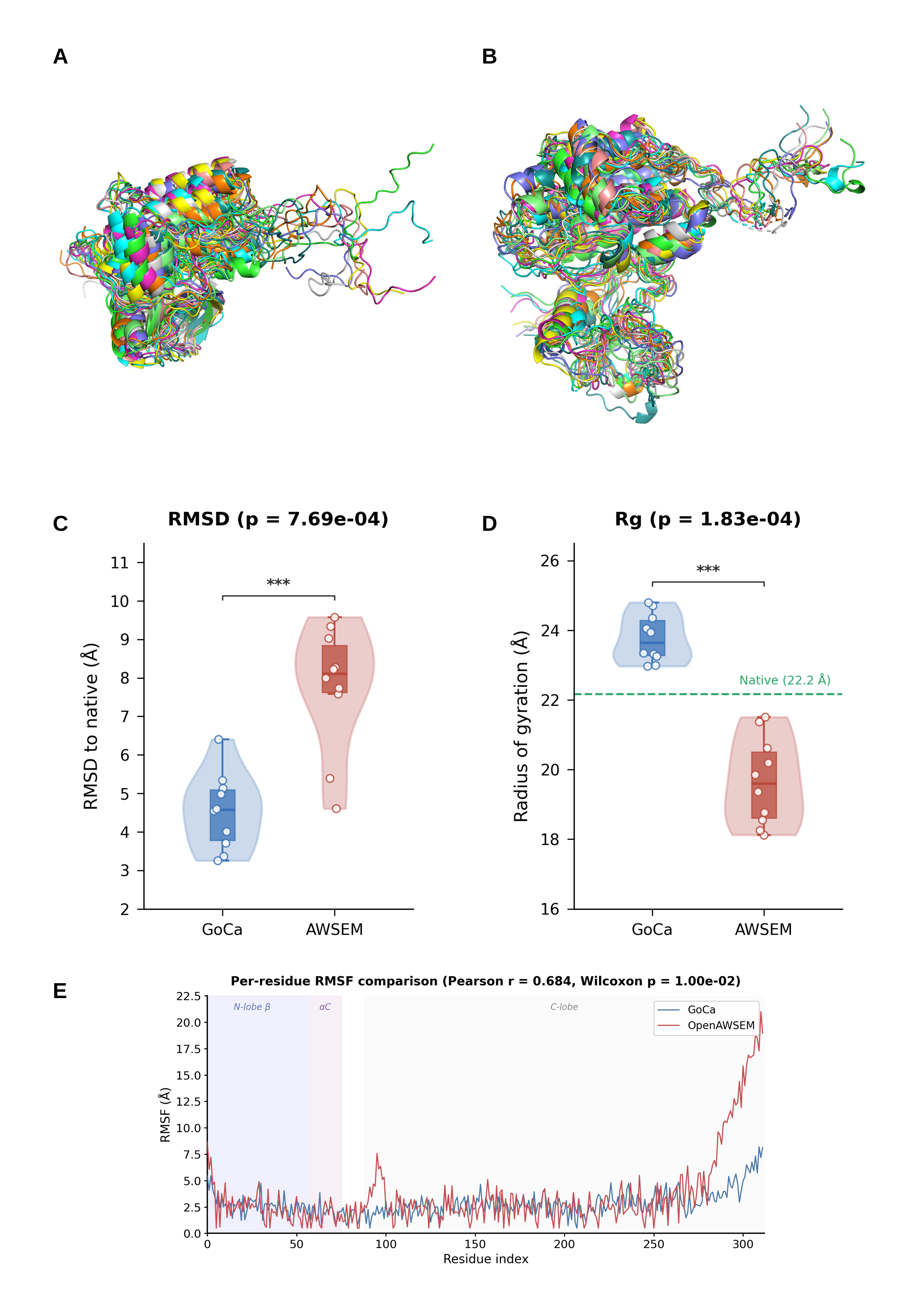}
    \caption{\textbf{Coarse-grained conformational sampling of the EGFR kinase domain by OpenAWSEM and GoCa.}
    (\textbf{A})~Superposition of 10 PULCHRA-reconstructed all-atom conformations from the OpenAWSEM ensemble, aligned to the 1M17 crystal structure.
    (\textbf{B})~Corresponding superposition for the GoCa ensemble.
    (\textbf{C})~C$\alpha$-RMSD to native structure: GoCa $4.54 \pm 0.93$~\AA\ versus AWSEM $7.78 \pm 1.53$~\AA\ ($P = 7.69 \times 10^{-4}$).
    (\textbf{D})~Radius of gyration: GoCa $R_\mathrm{g} = 23.77 \pm 0.65$~\AA, AWSEM $R_\mathrm{g} = 19.66 \pm 1.18$~\AA, native $R_\mathrm{g} = 22.2$~\AA\ (dashed line; $P = 1.83 \times 10^{-4}$).
    (\textbf{E})~Pairwise C$\alpha$-RMSD within each ensemble ($n = 45$ pairs): GoCa $4.28 \pm 0.88$~\AA\ versus AWSEM $5.67 \pm 1.60$~\AA\ ($P = 2.44 \times 10^{-5}$).
    (\textbf{F})~Per-residue RMSF profiles across 312 residues (Pearson $r = 0.685$); shaded regions denote the N-lobe $\beta$-sheet, $\alpha$C-helix, and C-lobe.
    Violin plots show individual data points with box-and-whisker summaries; $P$-values from two-sided Mann--Whitney $U$ tests; $^{***}P < 0.001$; $n = 10$ conformations per method.}
    \label{fig:cg-sampling}
\end{figure}

\noindent \textit{Coarse-grained conformational sampling and all-atom reconstruction of the EGFR kinase domain (E2E-Q1).}
To evaluate the agent's capacity for conformational sampling, a prerequisite for cryptic allosteric site discovery, we tasked it with performing dual coarse-grained (CG) molecular dynamics simulations of the EGFR kinase domain (PDB: 1M17, chain~A, 312 residues) using GoCa and OpenAWSEM force fields, followed by PULCHRA all-atom reconstruction. Superposition of the ten reconstructed ensembles onto the native crystal structure (Fig.~\ref{fig:cg-sampling}A,~B) revealed that both pipelines were executed end-to-end without manual intervention, producing physically plausible conformational ensembles suitable for downstream pocket analysis.

Quantitative assessment showed that GoCa, a native-topology G\={o} model, sampled conformations confined near the native basin (C$\alpha$-RMSD 3.25--6.40~\AA), whereas OpenAWSEM explored substantially broader conformational space (4.61--9.58~\AA; Mann--Whitney $U$, $P = 7.69 \times 10^{-4}$; Fig.~\ref{fig:cg-sampling}C). This broader sampling by OpenAWSEM is particularly relevant for cryptic site discovery, as transient pocket opening events typically require larger-amplitude backbone rearrangements that deviate significantly from the holo crystal structure.

The two force fields exhibited opposing trends in global compactness (Fig.~\ref{fig:cg-sampling}D). GoCa structures were mildly expanded relative to the native state (mean $R_\mathrm{g} = 23.8$~\AA\ versus native 22.2~\AA), while OpenAWSEM produced notably more compact conformations (mean $R_\mathrm{g} = 19.7$~\AA; $P = 1.83 \times 10^{-4}$). This compaction may reflect AWSEM's cooperative folding energy landscape favoring tight hydrophobic packing, potentially exposing surface cryptic pockets through differential domain rearrangement rather than global swelling. Pairwise C$\alpha$-RMSD among the ten sampled conformations further confirmed that OpenAWSEM generated greater structural diversity (mean 5.67~\AA\ versus 4.28~\AA; $P = 2.44 \times 10^{-5}$; Fig.~\ref{fig:cg-sampling}E), providing a broader ensemble for subsequent pocket detection.

Despite these quantitative differences, per-residue RMSF profiles from the two methods were moderately correlated (Pearson $r = 0.685$; Fig.~\ref{fig:cg-sampling}F), indicating consensus on the identity of flexible regions, including the N-/C-terminal tails, the $\alpha$C--$\beta$4 loop, and the activation segment, consistent with experimentally characterized EGFR dynamics implicated in allosteric regulation. The agent autonomously completed the full workflow encompassing structure cleaning, CG model preparation, MD execution, trajectory extraction, and all-atom reconstruction with appropriate quality controls, demonstrating its competence in conformational sampling tasks essential for structure-based drug discovery beyond the static crystallographic snapshot.

Beyond the scientific outcomes, the execution trace of E2E-Q1 (Fig.~\ref{fig:e2e-traces}A) reveals MolClaw's real-time adaptive capabilities under cascading tool failures. The agent encountered five distinct failures across the two simulation engines: GoCa returned empty outputs due to an incorrect tool name (the agent had invoked \texttt{goca\_pipeline} rather than the registered \texttt{run\_goca\_pipeline}), the OpenAWSEM frame extraction tool crashed 
with a Unicode encoding error, and two subsequent attempts to parse the GoCa trajectory via MDTraj failed due to topology mismatches. In each case, the hierarchical skill architecture guided recovery. Upon the initial GoCa failure, the L2 workflow skill directed the agent to proceed with OpenAWSEM first and defer the retry, a non-blocking strategy that preserved pipeline progress. When the OpenAWSEM extraction tool failed, the agent, following L3 Principle~14 (mandatory file collection) had already proactively downloaded all raw simulation outputs, enabling it to parse the trajectory PDB locally, map non-standard AWSEM residue names to canonical codes, and extract 10 frames without server-side tool support. For the GoCa retry, the agent autonomously enumerated all 74 tools registered on the server to diagnose the 
naming discrepancy, executed a dry-run validation before resubmission, and upon trajectory extraction failure, wrote a custom extractor that identified 45 frames in the XTC trajectory, discarded the first 20\% as equilibration, and sampled 10 evenly spaced conformations. The entire workflow, spanning seven tool types and four recovery episodes, completed autonomously, producing 20 all-atom structures that passed all 16 
pre-report data integrity checks.

\begin{figure}[p]
    \centering
    \includegraphics[width=0.9\linewidth,height=0.92\textheight,keepaspectratio]{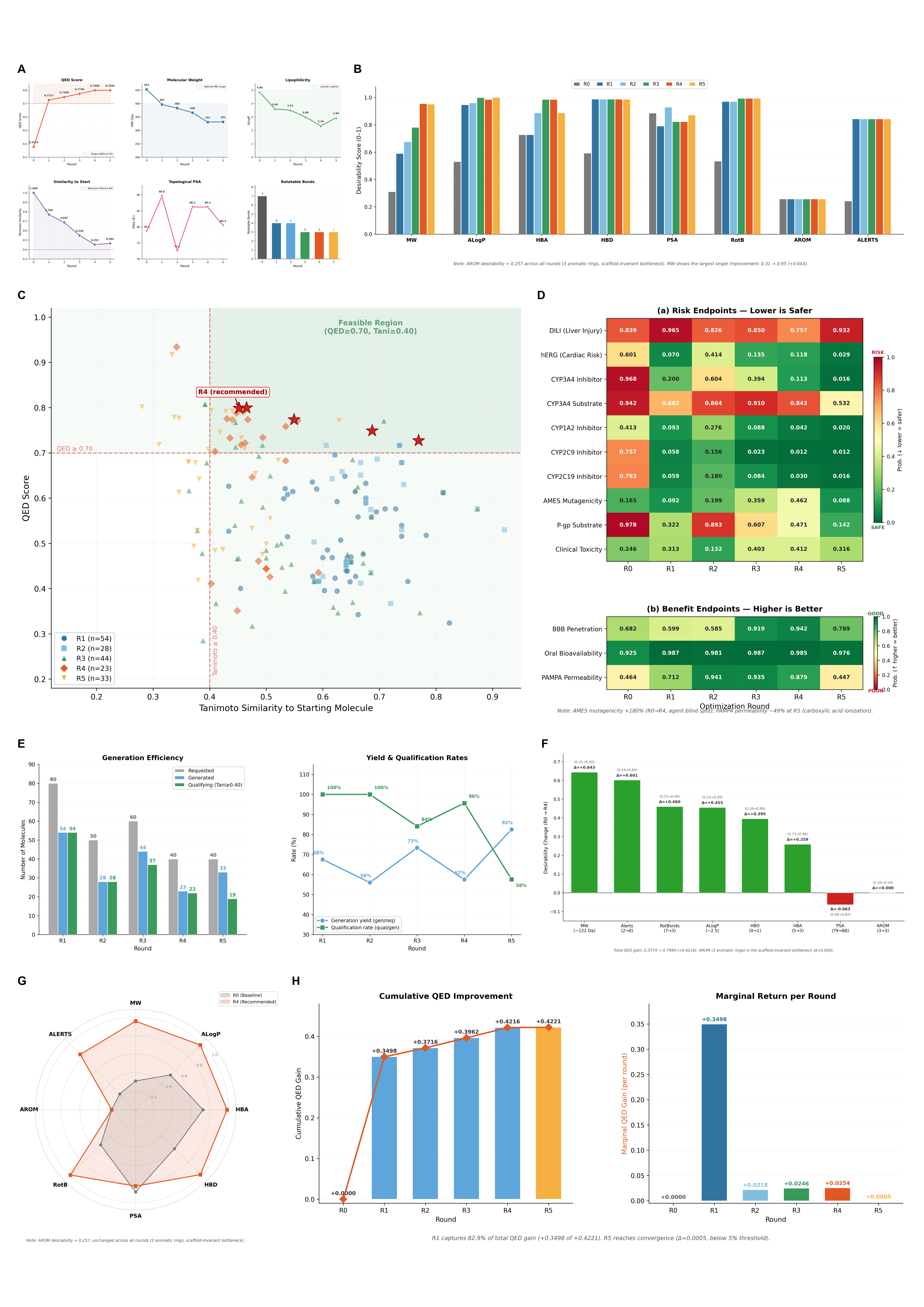}
    \captionsetup{font=footnotesize}
    \caption{\textbf{QED-driven iterative optimization of a triazolo-benzodiazepine scaffold by the AI agent.}
    (\textbf{A})~Multi-dimensional property trajectory across five optimization rounds: QED score (target $\geq$ 0.70), MW, ALogP, Tanimoto similarity (constraint $\geq$ 0.40), TPSA and rotatable bonds.
    (\textbf{B})~QED desirability decomposition by component and round (R0--R5).
    (\textbf{C})~QED--Tanimoto trade-off for all 182 molecules; red stars, selected best per round; green quadrant, feasible region.
    (\textbf{D})~ADMET profile evolution (top: risk endpoints; bottom: benefit endpoints).
    (\textbf{E})~Generation efficiency: molecules requested, generated and qualifying per round (left); yield and qualification rates (right).
    (\textbf{F})~QED component contribution waterfall (R0 $\to$ R4).
    (\textbf{G})~Radar chart comparing R0 and R4 desirability profiles.
    (\textbf{H})~Cumulative QED gain (left) and marginal gain per round (right).
    QED and Tanimoto computed via RDKit (Morgan fingerprint, radius = 2, 2{,}048 bits); ADMET by ADMET-AI; $n = 182$ molecules (54 + 28 + 44 + 23 + 33).}
    \label{fig:qed-opt}
\end{figure}

\noindent \textit{QED-driven iterative molecular optimization reveals scaffold-imposed ceilings and agent blind spots (E2E-Q2).}
We tasked the agent with iteratively optimizing the quantitative estimate of drug-likeness (QED) of a triazolo-benzodiazepine compound (MW = 452.9~Da, QED = 0.377) over five rounds, subject to dual constraints of QED $\geq$ 0.70 and Tanimoto similarity $\geq$ 0.40 to the starting molecule. The agent executed a complete Assess--Diagnose--Design--Verify loop at each round, using REINVENT4 mol2mol sampling for candidate generation, RDKit for property computation, and ADMET-AI for safety profiling. The full optimization trajectory is shown in Fig.~\ref{fig:qed-opt}.

The agent achieved the QED target in a single round (R1, QED = 0.727) by replacing the butyl ester side chain with a free carboxylic acid, simultaneously reducing molecular weight by 56~Da, removing three rotatable bonds, and clearing two Brenk structural alerts (Fig.~\ref{fig:qed-opt}A). This single modification captured 82.9\% of the total five-round QED improvement (Fig.~\ref{fig:qed-opt}H, right panel). Subsequent rounds yielded progressively smaller gains through targeted modifications: carboxylic acid to ethanol (R2, +0.022), methoxy group removal combined with alcohol-to-amide conversion (R3, +0.025), and chlorine removal from the pendant phenyl ring (R4, +0.025). By R5, the marginal gain fell to +0.0005, and the agent correctly declared convergence (Fig.~\ref{fig:qed-opt}H). The final recommended molecule (R4; QED = 0.799, Tanimoto = 0.452) represents a 112\% improvement over baseline.

To understand why QED plateaued near 0.80, we decomposed the score into its eight constituent desirability functions (Fig.~\ref{fig:qed-opt}B,~F). Six components improved dramatically from R0 to R4: MW (+0.643), structural alerts (+0.601), rotatable bonds (+0.460), ALogP (+0.455), HBD (+0.395) and HBA (+0.259). However, the aromatic ring desirability (AROM) remained fixed at 0.257 across all rounds, the triazolo-benzodiazepine scaffold inherently contains three aromatic rings, and this cannot be altered without breaking the core pharmacophore or violating the Tanimoto constraint. The radar comparison of R0 versus R4 profiles (Fig.~\ref{fig:qed-opt}G) visualizes this ceiling effect: the R4 polygon approaches a near-regular octagon except for a persistent indentation at the AROM axis. Using the weighted geometric mean formulation of QED, we derived a theoretical ceiling of 0.846; the achieved value of 0.799 represents 94--95\% of this bound.

The QED--Tanimoto scatter of all 182 generated molecules (Fig.~\ref{fig:qed-opt}C) reveals a fundamental trade-off: higher QED generally requires greater structural divergence. In rounds 3--5, the global highest-QED molecules (up to 0.934) fell below the Tanimoto threshold and were correctly excluded by the agent. Generation efficiency data (Fig.~\ref{fig:qed-opt}E) further quantify this tension: the fraction of molecules satisfying the Tanimoto constraint declined from 100\% (R1) to 57.6\% (R5) as the iterative seed progressively diverged from the starting structure.

Beyond QED optimization, the agent tracked multiple ADMET endpoints across rounds (Fig.~\ref{fig:qed-opt}D). CYP3A4 inhibition probability decreased from 0.968 to 0.113 ($-$88\%), hERG cardiac risk from 0.601 to 0.118 ($-$80\%), and aqueous solubility improved by 2.6 log units. The agent also demonstrated multi-objective judgement by selecting R4 over R5, correctly prioritizing the compound with the superior overall ADMET profile despite R5 having marginally higher QED. However, a critical blind spot emerged: AMES mutagenicity probability rose from 0.165 to 0.462 (+180\%) between R0 and R4, approaching the positive threshold, yet this deterioration was never flagged in the agent's reports. This omission likely reflects an attentional bias toward endpoints that were problematic at baseline while neglecting those that started in a safe range.

The execution trace of E2E-Q2 (Fig.~\ref{fig:e2e-traces}B) further illustrates the agent's real-time adaptive reasoning across the five-round optimization loop. In Round 1, the primary generation tool returned a JSON parsing error; the agent immediately executed its pre-planned fallback strategy, switching to an alternative tool without retrying, as this decision logic was encoded during Phase 0 planning. More subtly, the agent detected that the tool-computed baseline QED (0.377) diverged substantially from the task description's
estimate (0.50--0.55) and autonomously adopted the tool-derived value as ground truth throughout all subsequent rounds, following L3 Principle~13
(computation-first data authority). Strategy adaptation was explicitly conditioned on prior quantitative outcomes: when Round~2's conservative
high-similarity prior yielded only $\Delta\text{QED} = +0.022$, the agent diagnosed the gain as insufficient and reverted to medium-similarity
exploration for Round~3, subsequently escalating to matched molecular pair priors for surgical single-site modifications in Rounds~4--5. A critical
constraint-enforcement pattern emerged in later rounds: in Round~4 the highest-QED molecule (0.934) was rejected because its Tanimoto similarity
(0.34) fell below the 0.40 threshold, and in Round~5 the top two candidates were likewise excluded (42\% rejection rate), demonstrating that the agent consistently prioritized constraint compliance over objective maximization.
The three-tier verification system (Checkpoint~A after each tool call, Checkpoint~B per round, Checkpoint~C pre-report) audited all 19 key numerical
values against their source files with zero discrepancies.

\noindent \textit{Structure-guided iterative lead optimization of Erlotinib targeting EGFR (E2E-Q3).}
 
\begin{figure}[p]
    \centering
    \includegraphics[width=\linewidth,height=0.82\textheight,keepaspectratio]{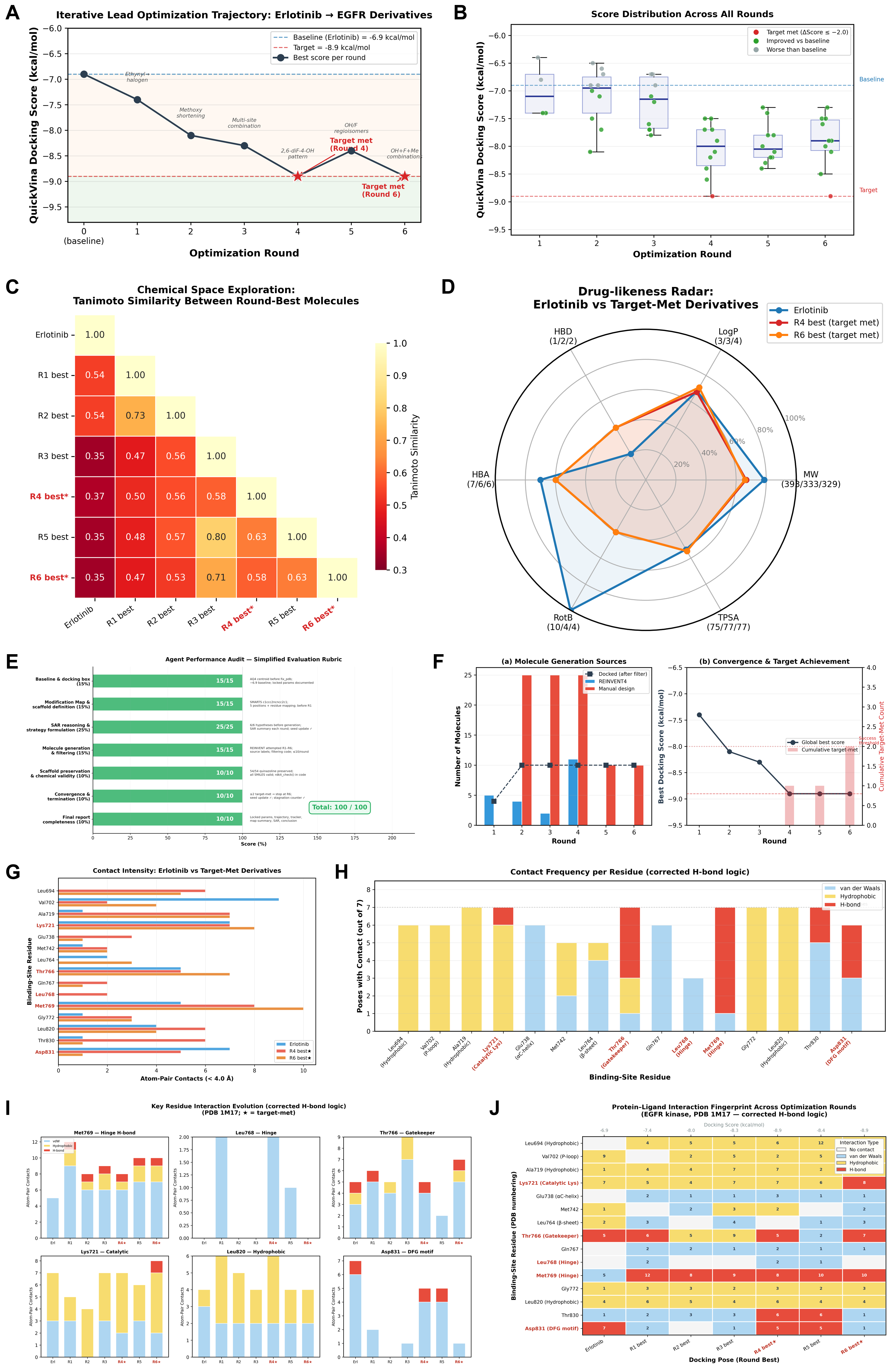}
    \captionsetup{font=footnotesize}
    \caption{\textbf{Comprehensive evaluation of AI-agent-driven iterative lead optimization of Erlotinib targeting the EGFR kinase domain.}
    (\textbf{A})~Optimization trajectory showing best QuickVina docking score per round; blue dashed line: Erlotinib baseline ($-6.9$~kcal/mol); red dashed line: $-8.9$~kcal/mol target.
    (\textbf{B})~Docking score distributions across Rounds~1--6 (box-and-strip plot, $n = 54$).
    (\textbf{C})~Tanimoto similarity heatmap between round-best molecules.
    (\textbf{D})~Drug-likeness radar plot comparing Erlotinib with two target-met derivatives.
    (\textbf{E})~Agent performance audit against seven-criterion rubric (100/100).
    (\textbf{F})~Molecule generation sources per round (left) and convergence curve (right).
    (\textbf{G})~Atom-pair contact counts ($< 4.0$~\AA) across 15 binding-site residues.
    (\textbf{H})~Contact frequency by interaction type across round-best poses.
    (\textbf{I})~Interaction evolution for six key residues across rounds.
    (\textbf{J})~Protein--ligand interaction fingerprint heatmap for all seven round-best poses.}
    \label{fig:q3-1}
\end{figure}

\noindent \textbf{Baseline establishment and locked docking parameters.}
The AI agent downloaded the EGFR kinase domain co-crystal structure (PDB: 1M17, chain~A, 2.6~\AA\ resolution) and extracted the AQ4 ligand centroid (22.014, 0.253, 52.794) from HETATM records prior to receptor preparation. After removing heterogens and waters and adding hydrogens at pH~7.0 via PDBFixer, the receptor was converted to PDBQT format and Erlotinib was docked using QuickVina2 with a $25 \times 25 \times 25$~\AA\ box centered on the AQ4 centroid. The resulting baseline docking score was $-6.9$~kcal/mol, establishing a target threshold of $-8.9$~kcal/mol ($\Delta\text{Score} \leq -2.0$). Erlotinib's computed molecular properties (MW = 393.4~Da, LogP = 3.41, HBD = 1, HBA = 7, rotatable bonds = 10, TPSA = 74.7~\AA$^2$) confirmed zero Lipinski violations and Veber compliance. These docking parameters were locked for all subsequent rounds (Fig.~\ref{fig:q3-1}A).
 
\noindent \textbf{Modification map and scaffold definition.}
Before entering the optimization loop, the agent defined a quinazoline core SMARTS pattern and enumerated five modifiable positions on the Erlotinib scaffold: (i) C6-methoxyethoxy, proximal to gatekeeper Thr766; (ii) C7-methoxyethoxy, extending toward solvent; (iii) C4-NH-aniline ring, projecting into the hydrophobic back pocket near Leu820; (iv) terminal ethynyl at the aniline meta-position; and (v) unsubstituted C2 on the quinazoline ring. Each position was mapped to nearby binding-site residues to guide rational modification hypotheses.
 
\begin{figure}[p]
    \centering
    \includegraphics[width=\linewidth,height=0.82\textheight,keepaspectratio]{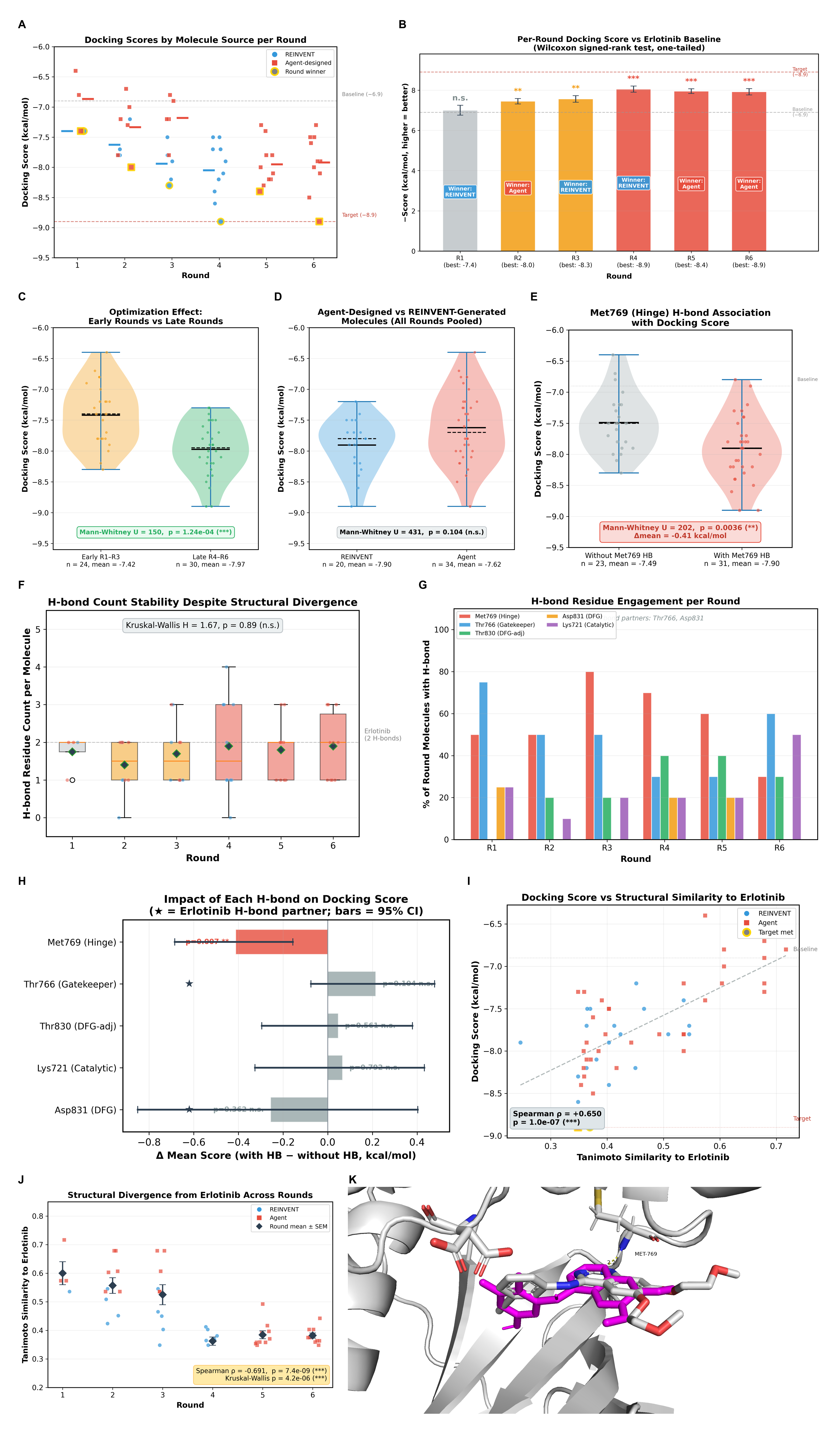}
    \captionsetup{font=footnotesize}
    \caption{\textbf{Statistical validation, source attribution, and interaction conservation analysis.}
    (\textbf{A})~Docking scores of all 54 molecules by source: REINVENT4 (blue, $n = 20$) and agent-designed (red, $n = 34$).
    (\textbf{B})~Per-round mean scores $\pm$ s.e.m.\ tested against baseline (Wilcoxon); R1 n.s., R2--R3 $^{**}$, R4--R6 $^{***}$.
    (\textbf{C})~Early (R1--R3) vs.\ late (R4--R6) violin plot ($p = 1.24 \times 10^{-4}$).
    (\textbf{D})~Agent vs.\ REINVENT violin plot ($p = 0.104$, n.s.).
    (\textbf{E})~H-bond residue count across rounds (Kruskal--Wallis $p = 0.89$, n.s.).
    (\textbf{F})~H-bond frequency for five key residues per round.
    (\textbf{G})~Met769 H-bond present vs.\ absent ($p = 0.0036$).
    (\textbf{H})~Forest plot of per-residue H-bond impact; only Met769 significant ($p = 0.007$).
    (\textbf{I})~Score vs.\ Tanimoto (Spearman $\rho = +0.650$, $p < 10^{-7}$).
    (\textbf{J})~Tanimoto across rounds ($\rho = -0.691$, $p = 7.4 \times 10^{-9}$).
    (\textbf{K})~3D rendering of R6 target-met compound in EGFR binding pocket with three-point H-bond network.
    $n = 54$ molecules; $^{**}P < 0.01$; $^{***}P < 0.001$.}
    \label{fig:q3-3}
\end{figure}

\noindent \textbf{Iterative optimization trajectory.}
The agent executed six rounds of the Strategize $\to$ Generate $\to$ Dock $\to$ Evaluate loop before achieving the convergence criterion ($\geq 2$ molecules with $\Delta\text{Score} \leq -2.0$). A total of 54 molecules were docked across all rounds (Fig.~\ref{fig:q3-1}A, Fig.~\ref{fig:q3-3}A). The best docking score improved monotonically from Rounds~0 through~4 ($-6.9 \to -7.4 \to -8.0 \to -8.3 \to -8.9$~kcal/mol), regressed to $-8.4$~kcal/mol in Round~5, and recovered to $-8.9$~kcal/mol in Round~6 (second target-met molecule), at which point the agent correctly terminated with success (Fig.~\ref{fig:q3-1}A). Assessed against the simplified seven-criterion evaluation rubric, the agent achieved a score of 100/100 (Fig.~\ref{fig:q3-1}E), with full credit in all categories: baseline establishment (15/15), modification map definition (15/15), SAR reasoning and strategy formulation (25/25), molecule generation and filtering (15/15), scaffold preservation and chemical validity (10/10), convergence and termination (10/10), and final report completeness (10/10).
 
\noindent \textbf{Source attribution and Agent--REINVENT complementarity.}
Across six rounds, 54 molecules were docked: 20 generated by REINVENT4 mol2mol and 34 designed by the AI agent through hardcoded analog lists in autonomously written pipeline recovery scripts (Fig.~\ref{fig:q3-3}A). The per-round breakdown of molecule sources and the cumulative convergence trajectory are shown in Fig.~\ref{fig:q3-1}F. All four pipeline scripts (v1--v4) were authored by the agent itself (Claude Sonnet 4.6 via Claude Code); no pipeline was pre-written or uploaded by the user, and the skills package contained zero target-specific hints (confirmed by grep for ``erlotinib'', ``EGFR'', ``1M17'' across all 58 skill files, zero matches). Round winners alternated between sources, REINVENT won Rounds~1, 3, and~4 while the agent won Rounds~2, 5, and~6, yielding a 3:3 tie. When pooled, no significant difference in docking scores was detected between Agent-designed and REINVENT-generated molecules (Mann--Whitney $U = 431$, $p = 0.104$; Fig.~\ref{fig:q3-3}D), suggesting that the two generation strategies operated at comparable quality levels while exploring complementary chemical space.
 
\noindent \textbf{Statistical validation of optimization efficacy.}
The iterative design loop produced statistically significant score improvements relative to the Erlotinib baseline. The docking score distributions across all six rounds are shown in Fig.~\ref{fig:q3-1}B. Round~1 did not differ significantly from baseline (Wilcoxon $p = 0.50$), but from Round~2 onward, all rounds achieved $p < 0.01$ (Fig.~\ref{fig:q3-3}B). When grouped by phase, late-round molecules (R4--R6, $n = 30$, mean $= -7.97$~kcal/mol) were significantly better than early-round molecules (R1--R3, $n = 24$, mean $= -7.42$~kcal/mol; Mann--Whitney $U = 150$, $p = 1.24 \times 10^{-4}$; Fig.~\ref{fig:q3-3}C). Notably, from Round~4 onward, 100\% of docked molecules (30/30) scored below the Erlotinib baseline, compared with only 79\% (19/24) in Rounds~1--3.
 
\noindent \textbf{Drug-likeness profile.}
Both target-met molecules exhibited improved drug-likeness relative to Erlotinib: reduced molecular weight (333.3 and 329.3~Da vs.\ 393.4~Da), lower rotatable bond count (4 vs.\ 10), maintained LogP (3.37 and 3.54 vs.\ 3.41), and increased hydrogen-bond donor capacity (HBD = 2 vs.\ 1), with zero Lipinski violations and full Veber compliance (Fig.~\ref{fig:q3-1}D). The quinazoline core was preserved in all 54 derivatives (100\% scaffold retention).
 
\noindent \textbf{Progressive structural divergence with preserved H-bond capacity.}
Tanimoto similarity to Erlotinib decreased significantly across rounds (Spearman $\rho = -0.691$, $p = 7.4 \times 10^{-9}$; Kruskal--Wallis $p = 4.2 \times 10^{-6}$), from a mean of 0.60 in Round~1 to 0.38 in Rounds~4--6 (Fig.~\ref{fig:q3-3}J). The pairwise Tanimoto heatmap among round-best molecules (Fig.~\ref{fig:q3-1}C) further illustrates progressive chemical space divergence across the optimization trajectory. This structural divergence correlated with improved docking scores (Spearman $\rho = +0.650$, $p = 1.0 \times 10^{-7}$; Fig.~\ref{fig:q3-3}I), confirming that productive optimization required moving substantially beyond the Erlotinib scaffold. Despite this divergence, the per-molecule hydrogen-bond residue count remained remarkably stable across all six rounds (Kruskal--Wallis $H = 1.67$, $p = 0.89$; Fig.~\ref{fig:q3-3}E), fluctuating between 1.4 and 1.9, close to the Erlotinib reference value of~2.
 
\noindent \textbf{H-bond residue exchange and Met769 as the key affinity driver.}
Residue-level analysis revealed a specific H-bond exchange pattern (Fig.~\ref{fig:q3-3}F), with the interaction evolution for six key binding-site residues shown in Fig.~\ref{fig:q3-1}I. Met769 (hinge region), which formed no hydrogen bond with Erlotinib in the baseline docking, was engaged by 30--80\% of derivatives across rounds, making it the most frequently contacted H-bond partner (31/54, 57\%). Conversely, Asp831 (DFG motif), one of Erlotinib's two H-bond partners, was retained by only 5/54 (9\%) of derivatives. Forest-plot analysis (Fig.~\ref{fig:q3-3}H) revealed that only the Met769 H-bond was significantly associated with improved binding ($\Delta\text{mean} = -0.41$~kcal/mol, $p = 0.007$; Fig.~\ref{fig:q3-3}G). The loss of Asp831 had no significant impact on score ($p = 0.362$). The optimization thus selectively acquired the most score-relevant H-bond (Met769) while dispensing with a non-contributing one (Asp831).
 
\begin{figure}[p]
    \centering
    \includegraphics[width=\linewidth,height=0.82\textheight,keepaspectratio]{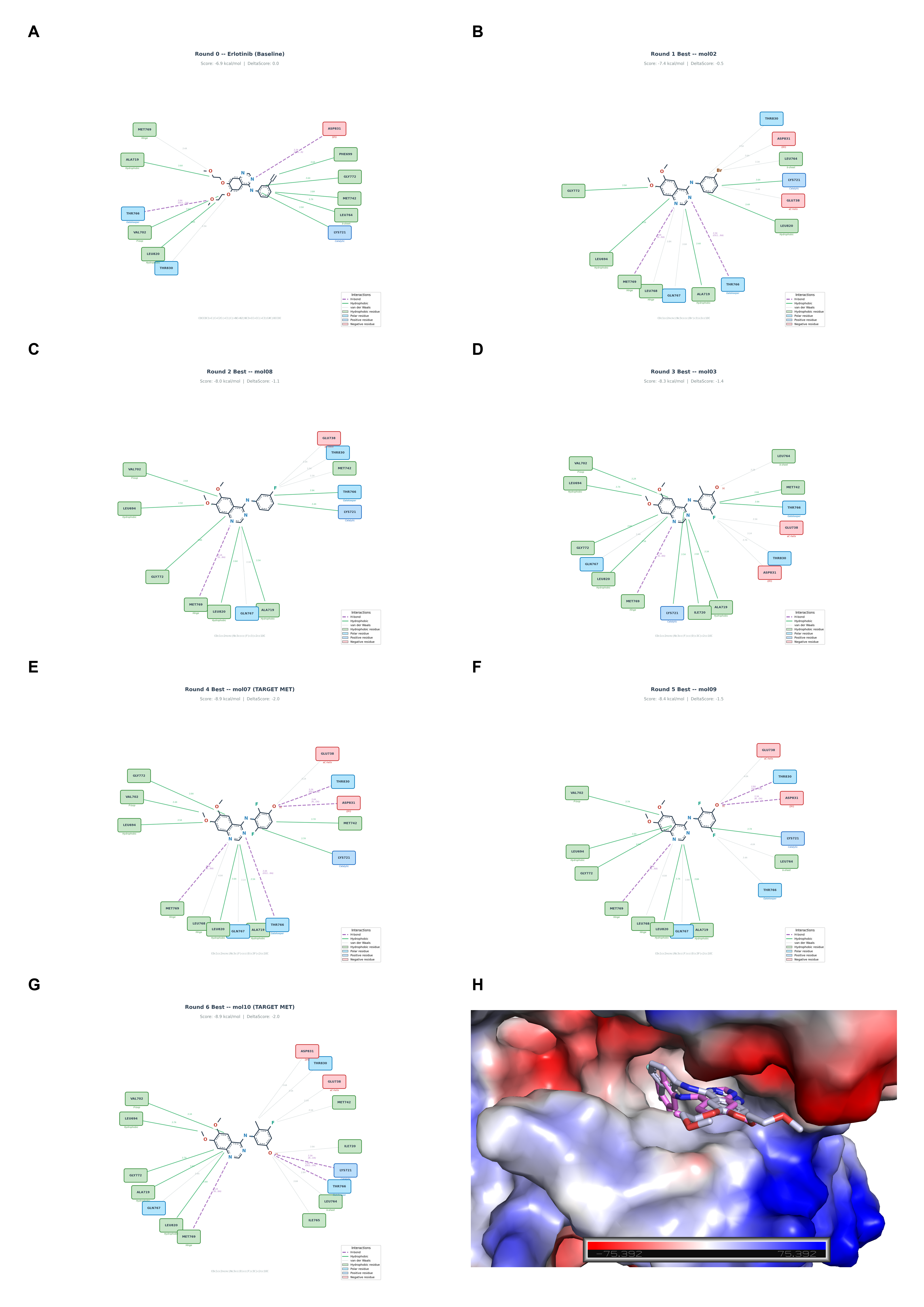}
    \captionsetup{font=footnotesize}
    \caption{\textbf{Schr\"{o}dinger-style 2D protein--ligand interaction diagrams and 3D pose overlay.}
    (\textbf{A})~Erlotinib baseline ($-6.9$~kcal/mol): two H-bonds (Thr766, Asp831) and eight hydrophobic contacts.
    (\textbf{B})~R1 best ($-7.4$): methoxy shortening + meta-Br; new Met769 H-bond (2.99~\AA).
    (\textbf{C})~R2 best ($-8.0$): Br$\to$F substitution; Met769 maintained.
    (\textbf{D})~R3 best ($-8.3$): F + OH + CH$_3$ on aniline.
    (\textbf{E})~R4 best ($-8.9$, target met): 2,6-diF-4-OH aniline; four simultaneous H-bonds (Met769, Thr766, Thr830, Asp831).
    (\textbf{F})~R5 best ($-8.4$): three H-bonds.
    (\textbf{G})~R6 best ($-8.9$, target met): distinct three-H-bond network (Lys721, Thr766, Met769).
    (\textbf{H})~3D rendering of representative poses within the EGFR ATP-binding pocket (electrostatic potential surface).}
    \label{fig:q3-2}
\end{figure}

\noindent \textbf{Protein-ligand interaction analysis and dual convergent binding modes.}
Distance-based interaction fingerprinting with corrected hydrogen-bond donor--acceptor logic was performed across all seven round-best poses (Fig.~\ref{fig:q3-1}J). Schr\"{o}dinger-style 2D protein--ligand interaction maps were generated for each round-best pose (Fig.~\ref{fig:q3-2}A--G). Quantitative atom-pair contact analysis across 15 binding-site residues (Fig.~\ref{fig:q3-1}G) and the interaction-type frequency distribution across round-best poses (Fig.~\ref{fig:q3-1}H) confirmed that hydrophobic contacts remained dominant throughout optimization. Erlotinib's binding was dominated by hydrophobic contacts with eight residues and two hydrogen bonds (Thr766, Asp831) (Fig.~\ref{fig:q3-2}A). The Round~4 target-met compound uniquely engaged four hydrogen bonds simultaneously, Met769 (3.12~\AA), Thr766 (3.36~\AA), Thr830 (3.18~\AA), and Asp831 (3.37~\AA), with Thr830 representing a genuinely novel contact absent in Erlotinib, while the Asp831 hydrogen bond was maintained but through a reversed donor--acceptor pairing (Fig.~\ref{fig:q3-2}E). The Round~6 target-met compound achieved the same $-8.9$~kcal/mol score via a different hydrogen-bond network, engaging Lys721 (3.32~\AA), Thr766 (3.25~\AA), and Met769 (3.15~\AA), without DFG-region contacts (Fig.~\ref{fig:q3-2}G, Fig.~\ref{fig:q3-3}K). These two distinct binding modes, a DFG-adjacent pathway (R4, 4 H-bonds) and a catalytic-lysine pathway (R6, 3 H-bonds), converge to the same docking score, demonstrating that the optimization landscape contains multiple degenerate minima accessible from the same starting scaffold.

The execution trace of E2E-Q3 (Fig.~\ref{fig:e2e-traces}C) exposes the full extent of autonomous adaptation required to complete this six-round
optimization campaign. Before the iterative loop could begin, baseline docking of Erlotinib failed four consecutive times across progressive box enlargements (25--50~\AA). Rather than terminating, the agent diagnosed the failure as a pipeline configuration issue and autonomously rewrote the execution script through four successive versions (v1--v4, totaling 163~KB of Python), ultimately recovering the baseline score of $-6.9$~kcal/mol. A second persistent failure affected interaction analysis: ProLIF crashed in every round due to valence and indexing errors in PDBQT-derived structures. The agent adapted by shifting to a score-only SAR strategy, formulating each round's modification hypothesis exclusively from quantitative docking results. For example, the Round~2 hypothesis explicitly cited Round~1's observation that methoxy shortening improved binding by 0.5~kcal/mol. A third failure emerged in Rounds~5--6, when REINVENT4 produced zero valid molecules as the seed diverged from its training distribution. The agent compensated by designing systematic regioisomer libraries manually: Round~5 explored 10 positional variants of the Round~4 scaffold's OH/F pattern, and Round~6 extended this to F/OH/CH$_3$ combinations, yielding the second target-met molecule. Throughout, the L2 workflow skill maintained a stagnation counter (reaching 2 of 3 before convergence) and enforced correct termination the moment the global tracker registered 2/2 target-met molecules.

\section*{Discussion}
 
The most instructive finding from our evaluation is not that MolClaw outperforms baselines, but \textit{where} it does and \textit{where} it does not. Across MolBench, MolClaw's hierarchical skills conferred zero benefit on property filtering (average gain: +0.0 percentage points across platforms), marginal improvement on molecule editing (+1.3~pp), but transformative gains on binding affinity comparison (+25.7~pp) and physicochemical optimization (2.0$\times$ delta increase). This monotonic relationship between skill-driven improvement and task complexity reveals a conceptual distinction that we believe has broad implications for scientific AI agent design: \textbf{tool access} and \textbf{workflow orchestration competence} are fundamentally different capabilities, and the latter constitutes the primary performance bottleneck for multi-step computational workflows.
 
Vanilla agent frameworks already possess tool access: they autonomously write and execute RDKit scripts, achieving $\geq$96\% accuracy on property filtering. Yet these same frameworks showed no improvement over standalone LLMs on binding affinity comparison (both at 51.4\%), as they cannot automatically invoke more sophisticated protein-ligand binding evaluation tools, let alone correctly orchestrate the full end-to-end execution workflow. Biomni's performance pattern provides a striking corroboration: it achieved 74\% on scriptable property filtering but fell to 24.3\% on binding affinity (below chance) and produced zero valid docking outputs \cite{huang2025biomni}, failing precisely on tasks requiring structured workflow orchestration. This pattern mirrors observations in software engineering agents, where ad hoc code generation suffices for isolated bug fixes but fails on multi-file refactoring requiring coordinated changes across components \cite{wang2023scientific}. MolClaw bridges this gap through its hierarchical skill architecture, where L2 workflow skills compose validated L1 tool modules into closed-loop pipelines with built-in quality gates, error recovery, and data handoff contracts between stages, encoding precisely the orchestration logic that ad hoc code generation cannot reliably reconstruct. MolBench, by deliberately stratifying tasks along this complexity axis, provides the first diagnostic framework for quantifying this distinction in drug discovery agents.
 
A persistent limitation of molecular optimization agents is the absence of structured convergence mechanisms \cite{bender2021artificial}. MolClaw addresses this through real-time collaboration across its three skill tiers: the L3 methodology layer (Principles 4--8) mandates that every optimization round follow an Assess$\to$Diagnose$\to$Design$\to$Verify loop with explicit quantitative justification; L2 workflow skills operationalize this mandate through COUNT GATE and MAPPING GATE checkpoints at each stage boundary, dynamically selecting and invoking the appropriate L1 modules based on task context; and L1 tool skills ensure that each computational step executes with validated parameters, standardized input/output schemas, and immediate post-execution quality controls. Crucially, these layers do not operate in isolation: L2 workflows cross-reference L3 principles at decision points (e.g., invoking Principle 3 to require cross-validation when docking results are ambiguous), while L1 skills propagate error signals upward, triggering L2 fallback protocols when tool calls fail. This hierarchical interaction, where L3 sets strategic constraints, L2 enforces procedural compliance and manages recovery, and L1 guarantees operational correctness, transforms the LLM from an unreliable ad hoc planner into a guideline-compliant executor of expert-validated protocols.
 
The E2E-Q2 optimization trajectory illustrates the practical consequence: a single targeted modification captured 82.9\% of the total QED improvement, marginal returns diminished monotonically from 0.350 (R1) to 0.0005 (R5), and the agent correctly declared convergence, demonstrating not just optimization capability but the ability to recognize when further optimization is unproductive. The scaffold-imposed QED ceiling analysis (theoretical maximum 0.846 due to locked AROM desirability at 0.257; achieved 0.799, representing 94--95\% of the bound) further validates that the agent's stopping decision was scientifically justified rather than premature. Equally important, when tool failures occurred during E2E-Q3, the agent's built-in fallback mechanisms enabled autonomous recovery: it authored four successive pipeline scripts (v1--v4), progressively diagnosing and resolving crashes from NumPy incompatibilities to f-string bugs, maintaining strategic coherence across disruptions. This self-repair capacity, governed by the L2 failure-recovery protocols, demonstrates that the hierarchical architecture provides resilience, not just accuracy. The E2E-Q3 results further revealed that the closed-loop design naturally accommodates complementary generation sources: REINVENT4 and agent-designed molecules contributed equally (3:3 round-winner tie, Mann--Whitney $P = 0.104$), with REINVENT excelling at creative recombination and the agent at systematic regioisomer scanning, a division of labor that emerged from the loop structure rather than explicit programming.
 
We note, however, that the current convergence criteria remain heuristic. Integrating acquisition functions from Bayesian optimization or surrogate-gradient methods into the iterative loop could provide stronger theoretical convergence guarantees, though this requires continuous parameterization of chemical space that remains an open challenge.
 
Most existing drug discovery agents operate on 1D molecular representations, treating 3D binding evaluation as a post hoc validation step rather than an integral optimization signal \cite{sadybekov2023computational}. MolClaw's E2E-Q3 task demonstrates a fundamentally different paradigm: docking is embedded within every iteration of the Strategize$\to$Generate$\to$Dock$\to$Evaluate loop, with locked parameters (box center, size, and scoring function held constant across all six rounds) ensuring that score differences between molecules are attributable to structural modifications rather than setup variability. This locked-parameter design, enforced by L2 workflow skills cross-referencing L3 Principle 19 (Docking Parameter Safeguards), exemplifies how hierarchical skill interaction converts a methodological best practice into an automatically enforced constraint.
 
The resulting systematic SAR analysis across 54 derivatives yielded two scientifically notable findings. First, residue-level interaction analysis identified Met769 (hinge region) as the sole statistically significant affinity driver ($\Delta$mean $= -0.41$~kcal/mol, $P = 0.007$), while the loss of Erlotinib's native Asp831 contact had no detectable impact, a selective H-bond exchange pattern consistent with established EGFR inhibitor SAR \cite{eberhardt2021autodock}. Second, two target-met molecules achieved the same $-8.9$~kcal/mol score through distinct hydrogen-bond networks (a DFG-adjacent pathway and a catalytic-lysine pathway), revealing degenerate minima in the optimization landscape that would be invisible to single-round evaluation.
 
These insights must be interpreted within the inherent limitations of docking-based assessment. Docking scores are ranking tools with limited correlation to experimental binding affinities \cite{sadybekov2023computational}, and the agent's ProLIF interaction analyses failed during real-time execution, the Met769 finding emerged from post hoc analysis rather than in-loop feedback. Bridging this gap by embedding higher-fidelity methods (MM-PBSA via gmx\_MMPBSA \cite{valdes2021gmx_mmpbsa}, or learned scoring functions such as EquiScore \cite{cao2024generic}) into the iterative loop represents a natural extension of MolClaw's structure-in-the-loop paradigm, trading computational cost for reliability at higher tiers of the screening funnel (L3 Principle 2).
 
While MolClaw demonstrates systematic advantages in structured workflows, three fundamental challenges remain unsolved. First, the AMES mutagenicity deterioration in E2E-Q2 (+180\%, from 0.165 to 0.462, approaching the positive threshold) was never flagged despite the agent tracking 13 ADMET endpoints \cite{swanson2024admet}. This reveals an attentional bias intrinsic to LLM-based optimization: endpoints problematic at baseline (CYP3A4 $= 0.968$, hERG $= 0.601$) attracted monitoring, while endpoints starting in a safe range escaped attention even as they gradually worsened. More broadly, the absence of explicit Pareto front management means that multi-objective trade-offs are resolved by implicit LLM judgment rather than structured decision logic, a limitation shared by all current drug discovery agents \cite{vamathevan2019applications}. We propose an ``unmonitored-endpoint alarm'' as an engineering solution: automatically flagging any ADMET endpoint that crosses a predefined deterioration threshold (e.g., absolute increase $>$0.15 or relative increase $>$100\%) regardless of the agent's explicit attention allocation.
 
Second, MolClaw's skill layer is static: the agent does not learn from prior executions. The recurring ``ethynyl fixation'' in E2E-Q3, where the agent hypothesized restoration of Erlotinib's terminal ethynyl in three separate rounds despite consistently negative evidence, demonstrates failure to inductively generalize from negative feedback. Fusing the deterministic workflow knowledge encoded in skills with probabilistic chemical intuition distilled from execution trajectories represents a critical frontier. The current architecture is deliberately designed to facilitate this fusion: because skills are structured documents decoupled from the LLM, they can serve as scaffolds for injecting learned knowledge without disrupting validated procedural logic.
 
Third, MolClaw's model- and platform-agnostic design carries a significant practical implication: the skill layer functions as a performance equalizer across heterogeneous LLM backends and agent runtimes, reducing the cross-platform performance gap (e.g., docking hit count disparity narrowed from 0.36 to 0.16 after skill introduction). This consistency across three LLMs (Claude Sonnet 4.6, Qwen 3.5, and Kimi 2.5) and two architecturally distinct platforms substantiates the claim that the skill layer, not the underlying model, is the primary driver of performance. One might argue that hierarchical skills represent sophisticated prompt engineering rather than methodological innovation. We contend that the distinction is meaningful: unlike prompts that suggest behaviors, MolClaw's skills prescribe verifiable execution protocols with formal data handoff contracts, quality gates, and failure recovery logic, creating an auditable chain from every reported value back to its computational source. This paradigm, converting implicit expert knowledge into composable, model-agnostic execution protocols, means MolClaw's capabilities will improve automatically as both LLMs and agent platforms advance \cite{gao2024empowering, feng2026internagent}.
 
Several limitations warrant explicit acknowledgment. All end-to-end results are purely computational; docking score improvements do not guarantee experimental affinity gains, and wet-laboratory validation remains essential before any translational claims can be made. The three E2E challenges, while detailed, represent a limited sample; expanding MolBench-E2E to diverse target classes (GPCRs, ion channels, proteases) and scaffold chemotypes is necessary to establish generalizability. The 70 skill documents required expert curation, and long-term scalability demands automated skill discovery from successful execution trajectories. Finally, context window constraints may limit simultaneous loading of L3, L2, and multiple L1 skills for the most complex workflows, motivating dynamic skill loading strategies.
 
Looking forward, we envision MolClaw's hierarchical skill paradigm as a generalizable design pattern for AI-driven scientific workflows. Any domain requiring multi-tool, multi-step, multi-objective orchestration, from materials discovery to synthetic biology to climate modeling, faces the same fundamental bottleneck that MolBench quantifies: tool access is increasingly commoditized, but the expertise to compose tools into scientifically valid workflows remains scarce and fragile when delegated to unstructured LLM reasoning \cite{wang2023scientific, boiko2023autonomous}. Encoding this expertise into composable, auditable, model-agnostic skill layers, rather than relying on ad hoc LLM planning or task-specific fine-tuning, offers a path toward AI agents that are not merely tool-users but genuine workflow-competent scientific partners.

\section*{Methods}

\subsection*{MolClaw Agent} 

As depicted in Figure~\ref{fig:agent}A, MolClaw is an autonomous computational drug discovery agent that integrates a unified tool infrastructure, a hierarchical skill ecosystem, and a gate-regulated execution framework to support end-to-end workflows from target preparation to lead optimization. By grounding every execution decision in domain-specific methodology and enabling progressive self-improvement through skill crystallization, MolClaw bridges the gap between general-purpose LLM reasoning and rigorous scientific practice.
\\[6pt]
\noindent \textbf{SCP Tool Infrastructure}

\noindent MolClaw standardizes over 30 specialized databases, software packages, and deep learning models into modular, interoperable services, enabling fully autonomous end-to-end computational drug discovery workflows without manual intervention in tool orchestration, format conversion, or parameter configuration.

For small-molecule evaluation, MolClaw integrates RDKit \cite{rdkit} for physicochemical property calculation, Open Babel for molecular format interconversion, ADMET-AI \cite{swanson2024admet} for high-throughput ADMET profiling, and DLEPS \cite{zhu2021prediction} for disease signature reversal prediction. These modules provide early-stage filtering and prioritization of candidate molecules based on drug-likeness, pharmacokinetics, and therapeutic relevance.

For molecular design, MolClaw separates small-molecule generation and biomolecular design into distinct functional modules. Small-molecule design tasks, including de novo generation, R-group replacement, and scaffold hopping, are supported via REINVENT 4 \cite{loeffler2024reinvent}, enabling flexible optimization of binding affinity, drug-likeness, and structural novelty. In parallel, protein and peptide design is supported through ProteinMPNN \cite{dauparas2022robust}, Chroma \cite{ingraham2023illuminating}, FoldX \cite{buss2018foldx}, and EvoBind \cite{bryant2022evobind}, enabling sequence–structure co-design and targeted binder generation for protein–ligand or protein–protein interactions.

For target structure preparation, MolClaw integrates ESMFold \cite{lin2023evolutionary} for rapid single-sequence structure prediction and Chai-1 \cite{chai2024chai} for protein–ligand and protein–protein complex modeling, with per-residue confidence metrics (pLDDT, pTM/ipTM) to identify low-confidence regions. Additional modules support multi-protein assembly simulation (GoCa \cite{walter2024structure}), all-atom reconstruction from C$\alpha$ traces (PULCHRA \cite{rotkiewicz2008fast}), protein ensemble modeling (BioEmu \cite{lewis2025scalable}), coarse-grained folding simulations (OpenAWSEM \cite{lu2021openawsem}), and automated visualization via PyMOL \cite{delano2002pymol}. Druggable binding site identification is performed using fpocket \cite{le2009fpocket} and P2Rank \cite{krivak2018p2rank}, with consensus-based selection improving robustness over single-method approaches. 

For binding pose prediction, MolClaw incorporates multiple complementary docking engines, including GPU-accelerated Vina-GPU 2.0 \cite{ding2023vina}, diffusion-based DiffDock \cite{corso2022diffdock}, and KarmaDock \cite{zhang2023efficient}, alongside HDOCK \cite{yan2020hdock} for protein–protein docking. This multi-method framework enables cross-validation of predicted binding poses, enhancing reliability in automated pipelines \cite{sadybekov2023computational}. Interaction characterization and scoring are performed using ProLIF \cite{bouysset2021prolif} for residue-level interaction fingerprints and PLIP \cite{salentin2015plip} for detailed annotation of hydrogen bonds, hydrophobic contacts, $\pi$-$\pi$ stacking, and salt bridges. Binding affinity prediction is further supported by EquiScore \cite{cao2024generic}, an equivariant graph neural network with physical priors, and Boltz-2 \cite{passaro2025boltz}, a foundation model for joint structure and affinity prediction.

For dynamic stability and thermodynamic validation, MolClaw integrates GROMACS \cite{abraham2015gromacs} and OpenMM \cite{eastman2017openmm} as molecular dynamics engines, with gmx\_MMPBSA \cite{valdes2021gmx_mmpbsa} enabling binding free energy estimation and per-residue energy decomposition.

MolClaw further integrates retrieval APIs from UniProt \cite{uniprot}, RCSB PDB \cite{rcsb_pdb}, AlphaFold DB \cite{alphafold_db}, PubChem \cite{pubchem}, and ChEMBL \cite{chembl}, forming a unified data acquisition pipeline for proteins and small molecules. All modules are uniformly encapsulated to ensure data traceability and eliminate manual intervention.

All tools are deployed on a GPU-accelerated cluster and exposed via a unified SCP infrastructure, delivering three core functionalities \cite{jiang2025scp}: (\romannumeral 1) standardized invocation via MCP-compliant input/output schemas; (\romannumeral 2) resource scheduling through dedicated job management for GPU-intensive tasks; (\romannumeral 3) access control with license-based authentication and per-IP request limits.
\\[6pt]
\noindent \textbf{Hierarchical Skill Ecosystem}

The MolClaw skill ecosystem comprises curated skill documents organized across three hierarchical layers, complemented by an external research layer and a self-expanding repository of auto-generated skills, described top-down from the most abstract strategic framework to concrete operational implementations.

The highest, \textbf{methodology-level (L3)} layer consists of a single comprehensive document encoding task-agnostic strategic meta-knowledge for drug discovery. Unlike lower layers that define specific operational steps, the L3 skill establishes 26 formal scientific principles across 8 chapters to govern agent decision-making, quality control, and result reporting. The first seven chapters address core execution competencies: systematic task decomposition prior to any tool invocation, progressive funnel-style screening workflows, mandatory cross-validation of critical results with a three-category epistemic labeling scheme for computed, predicted, and literature-derived values, standardized iterative optimization protocols governed by a three-question per-round discipline, rigorous data integrity safeguards including pre-reporting count verification and continuous self-audit checkpoints, structural biology awareness standards for residue numbering and docking box specification, and reproducible reporting conventions. Chapter 8, provided as a dedicated supplement, extends the framework with four additional principles governing autonomous scientific discovery and skill self-generation: structured capability landscape mapping and problem feasibility assessment, autonomous draft workflow authoring under strict quality and scope constraints, experiential crystallization of execution traces into persistent reusable skills, and dynamic capability boundary self-awareness. This document is loaded in full prior to workflow initiation, ensuring the agent adheres to robust scientific methodology even for novel tasks lacking dedicated workflow templates.

The intermediate, \textbf{workflow-level (L2)} layer comprises 14 task-specific documents providing end-to-end decision frameworks spanning the full early-stage drug discovery pipeline. Operating at the task level rather than at individual tool calls, these skills define tool selection criteria, method prioritization, quality gate enforcement, iteration rules, and failure recovery protocols for core use cases including target protein preparation, virtual screening, generative molecular design, binding affinity optimization, multi-target selectivity assessment, coarse-grained conformational sampling, and protein and peptide engineering. Three meta-workflows extend this layer beyond conventional task execution: a problem discovery and feasibility triage workflow that operationalizes structured capability landscape mapping for open-ended scientific tasks, a draft workflow authoring protocol that enables the agent to compose novel execution pipelines for tasks not covered by existing skills, and a skill crystallization workflow that formalizes the post-execution transformation of novel execution traces into persistent skill documents. Each workflow skill explicitly cross-references the relevant L3 principles, creating a hierarchical constraint chain that propagates strategic guidance to granular operational decisions.

The foundational, \textbf{tool-level (L1)} layer includes 60 fine-grained templates wrapping individual computational tools or functionally related tool families. A core design principle of this layer is task-specific decomposition: rather than monolithically wrapping entire software packages, multi-modal tools are partitioned into focused, single-purpose skills to minimize token consumption and interpretation error. The generative chemistry engine, for instance, is decomposed into seven task-specific skills covering distinct generation modes and reinforcement learning configurations, while molecular property calculation is split into granular modules for targeted descriptor sets. This layer also includes multi-step pipeline skills, closed-loop optimization skills that integrate the model's chemical reasoning with quantitative tool-derived feedback, and a skill template writer tool that provides standardized formatting scaffolds for auto-generated skill documents.

Beyond the three computational layers, MolClaw incorporates a \textbf{research-level (LR)} layer as a parallel, independent capability domain for external knowledge retrieval. This layer provides access to scientific literature, encyclopedic context, and web-accessible information through dedicated retrieval tools and a multi-source deep research synthesis workflow. The deep research workflow decomposes research questions into domain-specific sub-questions, executes searches across complementary sources with evidence hierarchy prioritization, and synthesizes findings into structured reports following drug discovery-specific guidelines. Critically, all LR-derived outputs carry mandatory provenance annotations and are governed by the computation-first principle: literature retrieval informs computational strategy and provides validation baselines but never substitutes for quantitative tool-derived results. Complementing these curated layers, an \textbf{auto-generated skill} repository accumulates workflow and tool-level knowledge crystallized from novel execution experiences. Skills enter this repository with an unvalidated confidence rating that is promoted through successive verified executions, and are indexed for retrieval during the planning phase of future tasks. This mechanism enables MolClaw to expand its effective skill coverage monotonically with accumulated execution history, without requiring manual expert curation for every encountered task variant.
\\[6pt]
\noindent \textbf{Agent Execution Framework}

A central design challenge in deploying LLM-based agents for multi-step scientific workflows is ensuring that domain expertise consistently shapes execution behavior without being overridden by general-purpose priors. MolClaw builds on Claude Code \cite{anthropic_claude_code} and OpenClaw \cite{openclaw} as the underlying execution substrate and coordinates them through a structured, gate-regulated execution framework. MolClaw coordinates tool use and reasoning through a structured, gate-regulated execution framework in which invariant procedural 
constraints, covering phase ordering, mandatory checkpoints, and planning prerequisites, are enforced at the infrastructure level. Complementarily, the L3 methodology skill encodes domain-specific principles for drug discovery that govern reasoning, planning, and quality evaluation. Together, they establish a strict epistemic ordering in which methodological knowledge is fully loaded prior to any execution decision, ensuring that tool selection, workflow design, and scientific judgment are grounded in domain principles rather than unconstrained model inference.

\textbf{Phase 0: Task Triage, Hierarchical Skill Reading, and Planning} is enforced as an absolute precondition by a hard gate. The agent first classifies the incoming task according to the coverage provided by existing curated workflow skills. Tasks with direct workflow coverage are routed to standard execution; composite tasks requiring the coordination of multiple workflow protocols must explicitly specify inter-phase data flow; open-ended discovery tasks are routed to a problem feasibility assessment procedure; and tasks without any existing workflow coverage trigger autonomous draft workflow authoring under strict quality constraints. Following triage, the agent performs hierarchical skill reading in a top-down sequence: the L3 methodology document is loaded in full before any workflow or tool skill is examined. Workflow-level skills are loaded selectively based on task relevance, and individual tool skills are deferred to on-demand loading during plan execution, which prevents context pollution from irrelevant tool documentation. An additional scan of the auto-generated skill index checks whether crystallized knowledge from prior executions applies to the current task; entries with sufficient validated confidence are loaded alongside expert-curated skills, enabling the system to benefit from its own execution history. Research needs are assessed at triage time through five structured signals, covering target biological context, structure-activity relationship context, seed molecule acquisition, novelty verification, and method selection guidance, which collectively determine whether the external research layer is activated. When activated, this layer provides literature retrieval, encyclopedic context, and web-accessible information that informs computational strategy without substituting for it; all externally retrieved values carry mandatory provenance annotations and are governed by a strict computation-first hierarchy. The execution plan produced in this phase integrates task decomposition, tool chain sequencing with explicit critical-path identification, fallback strategies, quality control checkpoint definitions, and research integration timing.

\textbf{Phase 1: Execution and Quality Assurance} governs sequential tool invocation under continuous verification. Each tool call is immediately followed by a sanity gate enforcing plausibility constraints such as valid score ranges, structural integrity, and non-empty outputs, with failures triggering automatic diagnosis, parameter adjustment, and retry. A subsequent cross-validation gate requires consistency across intermediate results against predefined thresholds. Iterative optimization tasks are further regulated by a three-question protocol specifying the target property, chemical strategy, and quantitative success criterion for each round, with explicit stopping conditions including convergence, target satisfaction, Pareto optimality, or resource exhaustion. A reactive literature search mechanism permits on-demand retrieval when unexpected execution results, such as anomalous scoring distributions, complete ADMET filter elimination, or stagnating optimization, warrant external contextualization. Symmetrically, a failure-triggered skill retrieval mechanism requires the agent to consult the crystallized skill index before designing any recovery strategy, propagating validated failure-recovery knowledge forward across sessions.

\textbf{Phase 2: Result Synthesis and Reporting} is initiated only after all quality control requirements are satisfied. A pre-report provenance audit maps each reported result to its originating file and validation step. Results are separated into three epistemically distinct categories: computed outputs, model interpretations, and literature-derived context. The mandatory post-execution self-assessment evaluates four crystallization triggers: successful completion of a novel tool composition workflow, successful resolution of a failure pattern absent from any existing skill documentation, a quantitatively validated protocol improvement over the standard approach, and identification of a systematic tool failure under reproducible input conditions. When any trigger is active, the agent executes a five-step crystallization procedure that transforms the execution trace into a persistent, reusable skill document through trace extraction, pattern abstraction with variable generalization, validation logic extraction, failure pattern formalization, and template-compliant formatting. The resulting skill is saved at the appropriate hierarchical level, with single-tool knowledge encoded as tool-level skills and multi-tool pipeline compositions encoded as workflow-level skills, and indexed with an initial unvalidated confidence rating that is promoted through successive verified executions. Critically, strategic methodology principles are never self-generated; their governance requires human expert validation, though the agent may propose candidate amendments as non-binding annotations alongside crystallized outputs.

Overall, this architecture treats reproducibility, methodological alignment, and progressive self-improvement as structural properties of the execution framework rather than emergent behaviors of the underlying model. The co-regulatory interplay between the primary procedural enforcement and the L3 skill's scientific principles, augmented by the auto-research and skill crystallization mechanisms, enables MolClaw to operate as a self-evolving scientific agent whose capability expands monotonically with accumulated execution experience while remaining bounded by explicit domain governance constraints.

\subsection*{MolBench Dataset}

\noindent \textbf{MolBench-MS}

\textit{Molecular property filtering.}
This task evaluates constraint satisfaction and structural recognition. Data are from the CARA lead optimization subset \cite{tian2024benchmarking}, grouped by Assay ChEMBL ID, retaining assays with $\geq 10$ unique SMILES after deduplication. For each assay, 10 SMILES were randomly sampled as candidates. We computed 11 RDKit descriptors per molecule: 6 physicochemical (molecular weight, MolLogP, hydrogen-bond donor/acceptor count, topological polar surface area, rotatable bond count) and 5 structural (ring count, aromatic ring count, fraction of sp3 carbons, heavy atom count, heteroatom count). Lipinski Rule of Five was applied as a baseline constraint, with 2–3 additional constraints per sample derived from quantiles of the candidate distribution. Iterative resampling ensured that 1–5 molecules satisfied all constraints.
Using the same assay pool, we further evaluated structural similarity reasoning. For each sample, 10 SMILES were sampled and canonicalized; one was selected as the query and the remaining 9 as candidates. Tanimoto similarity based on Morgan fingerprints defined ground truth. Two equivalent prompt variants were used: selecting the molecule with highest Tanimoto similarity, or the one sharing the most fingerprint fragments. 
Prompts listed rules and SMILES; outputs were evaluated via exact SMILES matching.

\textit{Binding affinity comparison.}
This task evaluates ranking of molecular binding affinity. Data were sourced from ACNet \cite{ren2023acnet}, containing experimental $K_i$ values. We selected 37 non-redundant targets via uniform sampling and one molecule pair per target. Each sample included a target and two molecules; models predicted either higher affinity (lower $K_i$) or lower affinity (higher $K_i$). Ground truth was determined by experimental $K_i$ values, and evaluation used exact SMILES matching.

\textit{Molecular docking screening.}
This task evaluates virtual screening using the CARA subset \cite{tian2024benchmarking}, retaining records with $\text{IC}_{50}$, $K_d$, or $K_i$. Assays were filtered to satisfy: (i) $\geq 60$ unique molecules; (ii) $\geq 6$ actives ($\text{pChEMBL} \geq 6$); (iii) $\geq 50$ inactives ($\text{pChEMBL} < 6$), yielding 25 assays. For each, 60 molecules (6–10 actives plus inactives) were sampled and shuffled. Models ranked candidates by predicted docking scores. Ground truth was defined by active molecules sorted by experimental $\text{pChEMBL}$ values. Performance was measured by Hits@3, defined as the average number of ground-truth active molecules retrieved in the model’s top 3 ranked results across all samples.
\\[6pt]
\noindent \textbf{MolBench-MO}

The MolBench-MO benchmark was derived from ChemCoTBench \cite{li2025beyond} by sampling tasks from two subtasks, molecular editing and physicochemical property optimization, while removing ambiguous or ill-defined entries.
The first subtask, functional group modification, requires models to add, delete, or substitute functional groups on a source molecule under strict constraints. Performance is measured by operational accuracy, defined as the proportion of outputs that fully satisfy all modification requirements.
The second subtask evaluates targeted optimization of drug-relevant physicochemical properties. Models must improve molecules with respect to three key metrics: QED, LogP, and LogS. Performance is assessed using two endpoints: (1) the absolute change (delta) in the target property and (2) the success rate in meeting predefined optimization criteria.

Tasks involving optimization toward three specific therapeutic targets were excluded. Preliminary experiments showed that LLMs primarily solved these tasks by recalling known medicinal chemistry strategies rather than performing genuine reasoning. Since MolBench-MO aims to evaluate reasoning ability rather than memorization, including such tasks would confound results and reduce the validity of cross-model comparisons.
\\[6pt]
\noindent \textbf{MolBench-E2E}

\noindent While MolBench-MS and MolBench-MO evaluate isolated molecular screening and single-metric optimization capabilities respectively, real-world computational drug discovery campaigns demand the autonomous orchestration of multi-tool, multi-phase workflows in which each step's output conditions the next step's input and execution logic. MolBench-E2E was designed to benchmark this long-horizon agentic competence: the ability to plan, execute, self-monitor, and adaptively revise complex scientific workflows that span 8--50+ sequential tool invocations across heterogeneous software ecosystems. The benchmark comprises three end-to-end challenges, each representing a distinct workflow archetype of early-stage structure-based drug discovery.

The first task (E2E-Q1) evaluates the agent's ability to perform coarse-grained conformational sampling using two force fields with fundamentally different theoretical foundations, GoCa (G\={o}-model-based C$\alpha$ coarse-grained dynamics) and OpenAWSEM (Associative-memory, Water-mediated, Structure and Energy Model), followed by all-atom reconstruction using PULCHRA. Starting from the EGFR kinase domain (PDB: 1M17), the agent must preprocess the crystal structure, prepare force-field-specific inputs for both simulation engines, execute independent coarse-grained molecular dynamics runs, extract representative conformational ensembles, and reconstruct full atomic detail from the reduced representations. This task tests the agent's capacity to manage format conversions and parameter configurations across fundamentally different simulation paradigms within a single coordinated pipeline.

The second task (E2E-Q2) assesses closed-loop, multi-round molecular property optimization driven by quantitative feedback. E2E-Q2 mandates a structured iterative protocol over up to five rounds: each round requires explicit property assessment, diagnostic reasoning identifying which structural features limit the target metric (QED), generation of candidate modifications with chemical rationale, and quantitative verification before seed selection. The task imposes convergence detection rules (early stopping upon target achievement; convergence declaration after consecutive rounds without improvement) and structural similarity constraints (Tanimoto $\geq$ 0.40), requiring the agent to balance exploitation of promising modifications against exploration of alternative strategies. The evaluation focus is on the quality of the Assess $\rightarrow$ Diagnose $\rightarrow$ Design $\rightarrow$ Verify reasoning loop across rounds, rather than raw optimization magnitude.

The third task (E2E-Q3), the most complex, requires the agent to integrate receptor preparation, binding site characterization, baseline docking, generative molecular design (via REINVENT 4, supplemented by manual analog design), molecular docking with locked parameters, drug-likeness filtering, and structure--activity relationship (SAR) interpretation within a multi-round optimization loop of up to 15 rounds. The agent must establish and maintain a persistent experimental context, docking box coordinates, baseline scores, a scaffold modification map linking modifiable positions to nearby binding-site residues, across all rounds. Termination is governed by dual criteria: success ($\geq 2$ molecules achieving a docking score improvement of $\geq 2$~kcal/mol relative to Erlotinib) or failure (round 15 reached without meeting the target). A mandatory strategy pivot rule requires the agent to shift to a different modifiable position or multi-site modification if the best score stagnates for three consecutive rounds. This task tests the full integration of tool orchestration, scientific reasoning, and autonomous decision-making that characterizes real medicinal chemistry campaigns.

All three tasks share a common evaluation protocol. Each provides the agent with a target protein specification, a starting molecule or structure, explicit success/failure criteria with quantitative thresholds, and a required output format including structured log files and a final summary report. Agent performance is assessed via task-specific rubrics that decompose each challenge into weighted criteria spanning: (1) correct tool selection and invocation sequence; (2) scientific validity of intermediate reasoning; (3) data integrity and parameter consistency across workflow steps; (4) appropriate termination behavior; and (5) completeness of output deliverables. Full task specifications, per-phase workflow details, and evaluation rubrics are provided in Supplementary Tables S5--S7.

\bibliography{ref.bib}

@article{sun202290,
  title={Why 90\% of clinical drug development fails and how to improve it?},
  author={Sun, Duxin and Gao, Wei and Hu, Hongxiang and Zhou, Simon},
  journal={Acta Pharmaceutica Sinica B},
  volume={12},
  number={7},
  pages={3049--3062},
  year={2022},
  publisher={Elsevier}
}

@article{wouters2020estimated,
  title={Estimated research and development investment needed to bring a new medicine to market, 2009-2018},
  author={Wouters, Olivier J and McKee, Martin and Luyten, Jeroen},
  journal={Jama},
  volume={323},
  number={9},
  pages={844--853},
  year={2020}
}

@article{sadybekov2023computational,
  title={Computational approaches streamlining drug discovery},
  author={Sadybekov, Anastasiia V and Katritch, Vsevolod},
  journal={Nature},
  volume={616},
  number={7958},
  pages={673--685},
  year={2023},
  publisher={Nature Publishing Group UK London}
}

@article{vamathevan2019applications,
  title={Applications of machine learning in drug discovery and development},
  author={Vamathevan, Jessica and Clark, Dominic and Czodrowski, Paul and Dunham, Ian and Ferran, Edgardo and Lee, George and Li, Bin and Madabhushi, Anant and Shah, Parantu and Spitzer, Michaela and others},
  journal={Nature reviews Drug discovery},
  volume={18},
  number={6},
  pages={463--477},
  year={2019},
  publisher={Nature Publishing Group UK London}
}

@article{lin2023evolutionary,
  title={Evolutionary-scale prediction of atomic-level protein structure with a language model},
  author={Lin, Zeming and Akin, Halil and Rao, Roshan and Hie, Brian and Zhu, Zhongkai and Lu, Wenting and Smetanin, Nikita and Verkuil, Robert and Kabeli, Ori and Shmueli, Yaniv and others},
  journal={Science},
  volume={379},
  number={6637},
  pages={1123--1130},
  year={2023},
  publisher={American Association for the Advancement of Science}
}

@article{jumper2021highly,
  title={Highly accurate protein structure prediction with AlphaFold},
  author={Jumper, John and Evans, Richard and Pritzel, Alexander and Green, Tim and Figurnov, Michael and Ronneberger, Olaf and Tunyasuvunakool, Kathryn and Bates, Russ and {\v{Z}}{\'\i}dek, Augustin and Potapenko, Anna and others},
  journal={nature},
  volume={596},
  number={7873},
  pages={583--589},
  year={2021},
  publisher={Nature Publishing Group UK London}
}

@article{le2009fpocket,
  title={Fpocket: an open source platform for ligand pocket detection},
  author={Le Guilloux, Vincent and Schmidtke, Peter and Tuffery, Pierre},
  journal={BMC bioinformatics},
  volume={10},
  number={1},
  pages={168},
  year={2009},
  publisher={Springer}
}

@article{krivak2018p2rank,
  title={P2Rank: machine learning based tool for rapid and accurate prediction of ligand binding sites from protein structure},
  author={Kriv{\'a}k, Radoslav and Hoksza, David},
  journal={Journal of cheminformatics},
  volume={10},
  number={1},
  pages={39},
  year={2018},
  publisher={Springer}
}

@article{eberhardt2021autodock,
  title={AutoDock Vina 1.2. 0: new docking methods, expanded force field, and python bindings},
  author={Eberhardt, Jerome and Santos-Martins, Diogo and Tillack, Andreas F and Forli, Stefano},
  journal={Journal of chemical information and modeling},
  volume={61},
  number={8},
  pages={3891--3898},
  year={2021},
  publisher={ACS Publications}
}

@article{friesner2004glide,
  title={Glide: a new approach for rapid, accurate docking and scoring. 1. Method and assessment of docking accuracy},
  author={Friesner, Richard A and Banks, Jay L and Murphy, Robert B and Halgren, Thomas A and Klicic, Jasna J and Mainz, Daniel T and Repasky, Matthew P and Knoll, Eric H and Shelley, Mee and Perry, Jason K and others},
  journal={Journal of medicinal chemistry},
  volume={47},
  number={7},
  pages={1739--1749},
  year={2004},
  publisher={ACS Publications}
}

@misc{rdkit,
  author       = {{RDKit Consortium}},
  title        = {RDKit: Open-Source Cheminformatics Software},
  howpublished = {\url{https://www.rdkit.org/}},
  year         = {2024}
}

@article{abraham2015gromacs,
  title={GROMACS: High performance molecular simulations through multi-level parallelism from laptops to supercomputers},
  author={Abraham, Mark James and Murtola, Teemu and Schulz, Roland and P{\'a}ll, Szil{\'a}rd and Smith, Jeremy C and Hess, Berk and Lindahl, Erik},
  journal={SoftwareX},
  volume={1},
  pages={19--25},
  year={2015},
  publisher={Elsevier}
}

@article{dharmasivam2025leading,
  title={Leading AI-driven drug discovery platforms: 2025 landscape and global outlook},
  author={Dharmasivam, Mahendiran and Kaya, Busra and Akinware, Adedoyin and Azad, Mahan Gholam and Richardson, Des R},
  journal={Pharmacological Reviews},
  pages={100102},
  year={2025},
  publisher={Elsevier}
}

@article{bender2021artificial,
  title={Artificial intelligence in drug discovery: what is realistic, what are illusions? Part 1: Ways to make an impact, and why we are not there yet},
  author={Bender, Andreas and Cort{\'e}s-Ciriano, Isidro},
  journal={Drug discovery today},
  volume={26},
  number={2},
  pages={511--524},
  year={2021},
  publisher={Elsevier}
}

@article{achiam2023gpt,
  title={Gpt-4 technical report},
  author={Achiam, Josh and Adler, Steven and Agarwal, Sandhini and Ahmad, Lama and Akkaya, Ilge and Aleman, Florencia Leoni and Almeida, Diogo and Altenschmidt, Janko and Altman, Sam and Anadkat, Shyamal and others},
  journal={arXiv preprint arXiv:2303.08774},
  year={2023}
}

@misc{anthropic_claude,
  author       = {{Anthropic}},
  title        = {The Claude model family},
  howpublished = {\url{https://www.anthropic.com/claude}},
  year         = {2024}
}

@article{team2024gemini,
  title={Gemini: A family of highly capable multimodal models. arXiv 2023},
  author={Team, Gemini and Anil, Rohan and Borgeaud, Sebastian and Alayrac, Jean-Baptiste and Yu, Jiahui and Soricut, Radu and Schalkwyk, Johan and Dai, Andrew M and Hauth, Anja and Millican, Katie and others},
  journal={arXiv preprint arXiv:2312.11805},
  year={2024}
}

@article{touvron2023llama,
  title={Llama: Open and efficient foundation language models},
  author={Touvron, Hugo and Lavril, Thibaut and Izacard, Gautier and Martinet, Xavier and Lachaux, Marie-Anne and Lacroix, Timoth{\'e}e and Rozi{\`e}re, Baptiste and Goyal, Naman and Hambro, Eric and Azhar, Faisal and others},
  journal={arXiv preprint arXiv:2302.13971},
  year={2023}
}

@article{bai2023qwen,
  title={Qwen technical report},
  author={Bai, Jinze and Bai, Shuai and Chu, Yunfei and Cui, Zeyu and Dang, Kai and Deng, Xiaodong and Fan, Yang and Ge, Wenbin and Han, Yu and Huang, Fei and others},
  journal={arXiv preprint arXiv:2309.16609},
  year={2023}
}

@article{schick2023toolformer,
  title={Toolformer: Language models can teach themselves to use tools},
  author={Schick, Timo and Dwivedi-Yu, Jane and Dess{\`\i}, Roberto and Raileanu, Roberta and Lomeli, Maria and Hambro, Eric and Zettlemoyer, Luke and Cancedda, Nicola and Scialom, Thomas},
  journal={Advances in neural information processing systems},
  volume={36},
  pages={68539--68551},
  year={2023}
}

@article{mialon2023augmented,
  title={Augmented language models: a survey},
  author={Mialon, Gr{\'e}goire and Dess{\`\i}, Roberto and Lomeli, Maria and Nalmpantis, Christoforos and Pasunuru, Ram and Raileanu, Roberta and Rozi{\`e}re, Baptiste and Schick, Timo and Dwivedi-Yu, Jane and Celikyilmaz, Asli and others},
  journal={arXiv preprint arXiv:2302.07842},
  year={2023}
}

@inproceedings{yao2022react,
  title={React: Synergizing reasoning and acting in language models},
  author={Yao, Shunyu and Zhao, Jeffrey and Yu, Dian and Du, Nan and Shafran, Izhak and Narasimhan, Karthik R and Cao, Yuan},
  booktitle={The eleventh international conference on learning representations},
  year={2022}
}

@article{wang2023scientific,
  title={Scientific discovery in the age of artificial intelligence},
  author={Wang, Hanchen and Fu, Tianfan and Du, Yuanqi and Gao, Wenhao and Huang, Kexin and Liu, Ziming and Chandak, Payal and Liu, Shengchao and Van Katwyk, Peter and Deac, Andreea and others},
  journal={Nature},
  volume={620},
  number={7972},
  pages={47--60},
  year={2023},
  publisher={Nature Publishing Group UK London}
}

@article{boiko2023autonomous,
  title={Autonomous chemical research with large language models},
  author={Boiko, Daniil A and MacKnight, Robert and Kline, Ben and Gomes, Gabe},
  journal={Nature},
  volume={624},
  number={7992},
  pages={570--578},
  year={2023},
  publisher={Nature Publishing Group UK London}
}

@article{gao2024empowering,
  title={Empowering biomedical discovery with AI agents},
  author={Gao, Shanghua and Fang, Ada and Huang, Yepeng and Giunchiglia, Valentina and Noori, Ayush and Schwarz, Jonathan Richard and Ektefaie, Yasha and Kondic, Jovana and Zitnik, Marinka},
  journal={Cell},
  volume={187},
  number={22},
  pages={6125--6151},
  year={2024},
  publisher={Elsevier}
}

@article{feng2026internagent,
  title={InternAgent-1.5: A Unified Agentic Framework for Long-Horizon Autonomous Scientific Discovery},
  author={Feng, Shiyang and Ma, Runmin and Yan, Xiangchao and Fan, Yue and Hu, Yusong and Huang, Songtao and Zhang, Shuaiyu and Cao, Zongsheng and Peng, Tianshuo and Yuan, Jiakang and others},
  journal={arXiv preprint arXiv:2602.08990},
  year={2026}
}

@misc{openclaw,
  author       = {{OpenClaw Contributors}},
  title        = {OpenClaw: an open-source framework for building tool-augmented LLM agents},
  howpublished = {\url{https://github.com/openclaw}},
  year         = {2025}
}

@misc{anthropic_claude_code,
  author       = {{Anthropic}},
  title        = {Claude Code: a command-line tool for agentic coding},
  howpublished = {\url{https://docs.anthropic.com/en/docs/claude-code}},
  year         = {2025}
}

@article{m2024augmenting,
  title={Augmenting large language models with chemistry tools},
  author={M. Bran, Andres and Cox, Sam and Schilter, Oliver and Baldassari, Carlo and White, Andrew D and Schwaller, Philippe},
  journal={Nature machine intelligence},
  volume={6},
  number={5},
  pages={525--535},
  year={2024},
  publisher={Nature Publishing Group UK London}
}

@article{huang2025biomni,
  title={Biomni: A general-purpose biomedical ai agent},
  author={Huang, Kexin and Zhang, Serena and Wang, Hanchen and Qu, Yuanhao and Lu, Yingzhou and Roohani, Yusuf and Li, Ryan and Qiu, Lin and Li, Gavin and Zhang, Junze and others},
  journal={biorxiv},
  year={2025}
}

@article{liu2024drugagent,
  title={Drugagent: Automating ai-aided drug discovery programming through llm multi-agent collaboration},
  author={Liu, Sizhe and Lu, Yizhou and Chen, Siyu and Hu, Xiyang and Zhao, Jieyu and Lu, Yingzhou and Zhao, Yue},
  journal={arXiv preprint arXiv:2411.15692},
  year={2024}
}

@article{gao2025txagent,
  title={TxAgent: an AI agent for therapeutic reasoning across a universe of tools},
  author={Gao, Shanghua and Zhu, Richard and Kong, Zhenglun and Noori, Ayush and Su, Xiaorui and Ginder, Curtis and Tsiligkaridis, Theodoros and Zitnik, Marinka},
  journal={arXiv preprint arXiv:2503.10970},
  year={2025}
}

@article{he2026democratising,
  title={Democratising real-world drug discovery through agentic AI},
  author={He, Jiazhen and Lai, Helen and Saigiridharan, Lakshidaa and Ghiandoni, Gian Marco and Jenei, Kinga and Gokalp, Umur and Nukovic, Ajsa and Engkvist, Ola and Janet, Jon Paul and Genheden, Samuel},
  journal={Drug Discovery Today},
  pages={104605},
  year={2026},
  publisher={Elsevier}
}

@article{vichentijevikj2026prompt,
  title={Prompt-to-pill: Multi-Agent drug discovery and clinical simulation pipeline},
  author={Vichentijevikj, Ivana and Mishev, Kostadin and Simjanoska Misheva, Monika},
  journal={Bioinformatics Advances},
  volume={6},
  number={1},
  pages={vbaf323},
  year={2026},
  publisher={Oxford University Press}
}

@article{pan2025frogent,
  title={Frogent: An end-to-end full-process drug design agent},
  author={Pan, Qihua and Xu, Dong and Yao, Jenna Xinyi and Ma, Lijia and Zhu, Zexuan and Ji, Junkai},
  journal={arXiv preprint arXiv:2508.10760},
  year={2025}
}

@article{cao2026mozi,
  title={Mozi: Governed Autonomy for Drug Discovery LLM Agents},
  author={Cao, He and Liu, Siyu and Zhang, Fan and Liu, Zijing and Li, Hao and Feng, Bin and Bai, Shengyuan and Chen, Leqing and Xie, Kai and Li, Yu},
  journal={arXiv preprint arXiv:2603.03655},
  year={2026}
}

@inproceedings{averly2025liddia,
  title={Liddia: Language-based intelligent drug discovery agent},
  author={Averly, Reza and Baker, Frazier N and Watson, Ian A and Ning, Xia},
  booktitle={Proceedings of the 2025 Conference on Empirical Methods in Natural Language Processing},
  pages={12015--12039},
  year={2025}
}

@article{jiang2025scp,
  title={SCP: Accelerating Discovery with a Global Web of Autonomous Scientific Agents},
  author={Jiang, Yankai and Lou, Wenjie and Wang, Lilong and Tang, Zhenyu and Feng, Shiyang and Lu, Jiaxuan and Sun, Haoran and Pan, Yaning and Gu, Shuang and Su, Haoyang and others},
  journal={arXiv preprint arXiv:2512.24189},
  year={2025}
}

@article{chai2024chai,
  title={Chai-1: Decoding the molecular interactions of life},
  author={Chai Discovery team and Boitreaud, Jacques and Dent, Jack and McPartlon, Matthew and Meier, Joshua and Reis, Vinicius and Rogozhonikov, Alex and Wu, Kevin},
  journal={BioRxiv},
  pages={2024--10},
  year={2024},
  publisher={Cold Spring Harbor Laboratory}
}

@article{ding2023vina,
  title={Vina-GPU 2.0: further accelerating AutoDock Vina and its derivatives with graphics processing units},
  author={Ding, Ji and Tang, Shidi and Mei, Zheming and Wang, Lingyue and Huang, Qinqin and Hu, Haifeng and Ling, Ming and Wu, Jiansheng},
  journal={Journal of chemical information and modeling},
  volume={63},
  number={7},
  pages={1982--1998},
  year={2023},
  publisher={ACS Publications}
}

@article{corso2022diffdock,
  title={Diffdock: Diffusion steps, twists, and turns for molecular docking},
  author={Corso, Gabriele and St{\"a}rk, Hannes and Jing, Bowen and Barzilay, Regina and Jaakkola, Tommi},
  journal={arXiv preprint arXiv:2210.01776},
  year={2022}
}

@article{zhang2023efficient,
  title={Efficient and accurate large library ligand docking with KarmaDock},
  author={Zhang, Xujun and Zhang, Odin and Shen, Chao and Qu, Wanglin and Chen, Shicheng and Cao, Hanqun and Kang, Yu and Wang, Zhe and Wang, Ercheng and Zhang, Jintu and others},
  journal={Nature Computational Science},
  volume={3},
  number={9},
  pages={789--804},
  year={2023},
  publisher={Nature Publishing Group US New York}
}

@article{bouysset2021prolif,
  title={ProLIF: a library to encode molecular interactions as fingerprints},
  author={Bouysset, C{\'e}dric and Fiorucci, S{\'e}bastien},
  journal={Journal of cheminformatics},
  volume={13},
  number={1},
  pages={72},
  year={2021},
  publisher={Springer}
}

@article{cao2024generic,
  title={Generic protein--ligand interaction scoring by integrating physical prior knowledge and data augmentation modelling},
  author={Cao, Duanhua and Chen, Geng and Jiang, Jiaxin and Yu, Jie and Zhang, Runze and Chen, Mingan and Zhang, Wei and Chen, Lifan and Zhong, Feisheng and Zhang, Yingying and others},
  journal={Nature Machine Intelligence},
  volume={6},
  number={6},
  pages={688--700},
  year={2024},
  publisher={Nature Publishing Group UK London}
}

@article{passaro2025boltz,
  title={Boltz-2: Towards accurate and efficient binding affinity prediction},
  author={Passaro, Saro and Corso, Gabriele and Wohlwend, Jeremy and Reveiz, Mateo and Thaler, Stephan and Somnath, Vignesh Ram and Getz, Noah and Portnoi, Tally and Roy, Julien and Stark, Hannes and others},
  journal={BioRxiv},
  year={2025}
}

@article{eastman2017openmm,
  title={OpenMM 7: Rapid development of high performance algorithms for molecular dynamics},
  author={Eastman, Peter and Swails, Jason and Chodera, John D and McGibbon, Robert T and Zhao, Yutong and Beauchamp, Kyle A and Wang, Lee-Ping and Simmonett, Andrew C and Harrigan, Matthew P and Stern, Chaya D and others},
  journal={PLoS computational biology},
  volume={13},
  number={7},
  pages={e1005659},
  year={2017},
  publisher={Public Library of Science San Francisco, CA USA}
}

@article{loeffler2024reinvent,
  title={Reinvent 4: Modern AI--driven generative molecule design},
  author={Loeffler, Hannes H and He, Jiazhen and Tibo, Alessandro and Janet, Jon Paul and Voronov, Alexey and Mervin, Lewis H and Engkvist, Ola},
  journal={Journal of Cheminformatics},
  volume={16},
  number={1},
  pages={20},
  year={2024},
  publisher={Springer}
}

@article{valdes2021gmx_mmpbsa,
  title={gmx\_MMPBSA: a new tool to perform end-state free energy calculations with GROMACS},
  author={Vald{\'e}s-Tresanco, Mario S and Vald{\'e}s-Tresanco, Mario E and Valiente, Pedro A and Moreno, Ernesto},
  journal={Journal of chemical theory and computation},
  volume={17},
  number={10},
  pages={6281--6291},
  year={2021},
  publisher={ACS Publications}
}

@article{swanson2024admet,
  title={ADMET-AI: a machine learning ADMET platform for evaluation of large-scale chemical libraries},
  author={Swanson, Kyle and Walther, Parker and Leitz, Jeremy and Mukherjee, Souhrid and Wu, Joseph C and Shivnaraine, Rabindra V and Zou, James},
  journal={Bioinformatics},
  volume={40},
  number={7},
  pages={btae416},
  year={2024},
  publisher={Oxford University Press}
}

@article{dauparas2022robust,
  title={Robust deep learning--based protein sequence design using ProteinMPNN},
  author={Dauparas, Justas and Anishchenko, Ivan and Bennett, Nathaniel and Bai, Hua and Ragotte, Robert J and Milles, Lukas F and Wicky, Basile IM and Courbet, Alexis and de Haas, Rob J and Bethel, Neville and others},
  journal={Science},
  volume={378},
  number={6615},
  pages={49--56},
  year={2022},
  publisher={American Association for the Advancement of Science}
}

@article{salentin2015plip,
  title={PLIP: fully automated protein--ligand interaction profiler},
  author={Salentin, Sebastian and Schreiber, Sven and Haupt, V Joachim and Adasme, Melissa F and Schroeder, Michael},
  journal={Nucleic acids research},
  volume={43},
  number={W1},
  pages={W443--W447},
  year={2015},
  publisher={Oxford University Press}
}

@article{ren2023acnet,
  title={Activity cliff prediction: Dataset and benchmark},
  author={Zhang, Ziqiao and Zhao, Bangyi and Xie, Ailin and Bian, Yatao and Zhou, Shuigeng},
  journal={arXiv preprint arXiv:2302.07541},
  year={2023}
}

@article{tian2024benchmarking,
  title={Benchmarking compound activity prediction for real-world drug discovery applications},
  author={Tian, Tingzhong and Li, Shuya and Zhang, Ziting and Chen, Lin and Zou, Ziheng and Zhao, Dan and Zeng, Jianyang},
  journal={Communications Chemistry},
  volume={7},
  number={1},
  pages={127},
  year={2024},
  publisher={Nature Publishing Group UK London}
}

@article{li2025beyond,
  title={Beyond Chemical QA: Evaluating LLM's Chemical Reasoning with Modular Chemical Operations},
  author={Li, Hao and Cao, He and Feng, Bin and Shao, Yanjun and Tang, Xiangru and Yan, Zhiyuan and Yuan, Li and Tian, Yonghong and Li, Yu},
  journal={arXiv preprint arXiv:2505.21318},
  year={2025}
}

@article{walter2024structure,
  title={Structure-based protein assembly simulations including various binding sites and conformations},
  author={Walter, Luis J and Quoika, Patrick K and Zacharias, Martin},
  journal={Journal of Chemical Information and Modeling},
  volume={64},
  number={8},
  pages={3465--3476},
  year={2024},
  publisher={ACS Publications}
}

@article{rotkiewicz2008fast,
  title={Fast procedure for reconstruction of full-atom protein models from reduced representations},
  author={Rotkiewicz, Piotr and Skolnick, Jeffrey},
  journal={Journal of computational chemistry},
  volume={29},
  number={9},
  pages={1460--1465},
  year={2008},
  publisher={Wiley Online Library}
}

@article{lewis2025scalable,
  title={Scalable emulation of protein equilibrium ensembles with generative deep learning},
  author={Lewis, Sarah and Hempel, Tim and Jim{\'e}nez-Luna, Jos{\'e} and Gastegger, Michael and Xie, Yu and Foong, Andrew YK and Satorras, Victor Garc{\'\i}a and Abdin, Osama and Veeling, Bastiaan S and Zaporozhets, Iryna and others},
  journal={Science},
  volume={389},
  number={6761},
  pages={eadv9817},
  year={2025},
  publisher={American Association for the Advancement of Science}
}

@article{lu2021openawsem,
  title={OpenAWSEM with Open3SPN2: A fast, flexible, and accessible framework for large-scale coarse-grained biomolecular simulations},
  author={Lu, Wei and Bueno, Carlos and Schafer, Nicholas P and Moller, Joshua and Jin, Shikai and Chen, Xun and Chen, Mingchen and Gu, Xinyu and Davtyan, Aram and de Pablo, Juan J and others},
  journal={PLoS computational biology},
  volume={17},
  number={2},
  pages={e1008308},
  year={2021},
  publisher={Public Library of Science San Francisco, CA USA}
}

@article{delano2002pymol,
  title={Pymol: An open-source molecular graphics tool},
  author={DeLano, Warren L and others},
  journal={CCP4 Newsl. protein crystallogr},
  volume={40},
  number={1},
  pages={82--92},
  year={2002}
}

@article{ingraham2023illuminating,
  title={Illuminating protein space with a programmable generative model},
  author={Ingraham, John B and Baranov, Max and Costello, Zak and Barber, Karl W and Wang, Wujie and Ismail, Ahmed and Frappier, Vincent and Lord, Dana M and Ng-Thow-Hing, Christopher and Van Vlack, Erik R and others},
  journal={Nature},
  volume={623},
  number={7989},
  pages={1070--1078},
  year={2023},
  publisher={Nature Publishing Group UK London}
}

@article{bryant2022evobind,
  title={EvoBind: in silico directed evolution of peptide binders with AlphaFold},
  author={Bryant, Patrick and Elofsson, Arne},
  journal={bioRxiv},
  pages={2022--07},
  year={2022},
  publisher={Cold Spring Harbor Laboratory}
}

@article{yan2020hdock,
  title={The HDOCK server for integrated protein--protein docking},
  author={Yan, Yumeng and Tao, Huanyu and He, Jiahua and Huang, Sheng-You},
  journal={Nature protocols},
  volume={15},
  number={5},
  pages={1829--1852},
  year={2020},
  publisher={Nature Publishing Group UK London}
}

@article{zhu2021prediction,
  title={Prediction of drug efficacy from transcriptional profiles with deep learning},
  author={Zhu, Jie and Wang, Jingxiang and Wang, Xin and Gao, Mingjing and Guo, Bingbing and Gao, Miaomiao and Liu, Jiarui and Yu, Yanqiu and Wang, Liang and Kong, Weikaixin and others},
  journal={Nature biotechnology},
  volume={39},
  number={11},
  pages={1444--1452},
  year={2021},
  publisher={Nature Publishing Group US New York}
}

@misc{uniprot,
  author       = {{The UniProt Consortium}},
  title        = {UniProt: The Universal Protein Knowledgebase},
  howpublished = {\url{https://www.uniprot.org/}},
  year         = {2025}
}

@misc{rcsb_pdb,
  author       = {{RCSB PDB Consortium}},
  title        = {RCSB PDB: Research Collaboratory for Structural Bioinformatics Protein Data Bank},
  howpublished = {\url{https://www.rcsb.org/}},
  year         = {2024}
}

@misc{alphafold_db,
  author       = {{AlphaFold Database Consortium}},
  title        = {AlphaFold DB: Open Repository of Protein Structure Predictions},
  howpublished = {\url{https://alphafold.ebi.ac.uk/}},
  year         = {2024}
}

@misc{pubchem,
  author       = {{PubChem Consortium}},
  title        = {PubChem: Open Chemistry Database},
  howpublished = {\url{https://pubchem.ncbi.nlm.nih.gov/}},
  year         = {2025}
}

@misc{chembl,
  author       = {{ChEMBL Consortium}},
  title        = {ChEMBL: The Global Bioactivity Database for Drug Discovery},
  howpublished = {\url{https://www.ebi.ac.uk/chembl/}},
  year         = {2024}
}

@article{bai2025intern,
  title={Intern-s1: A scientific multimodal foundation model},
  author={Bai, Lei and Cai, Zhongrui and Cao, Yuhang and Cao, Maosong and Cao, Weihan and Chen, Chiyu and Chen, Haojiong and Chen, Kai and Chen, Pengcheng and Chen, Ying and others},
  journal={arXiv preprint arXiv:2508.15763},
  year={2025}
}

@article{buss2018foldx,
  title={FoldX as protein engineering tool: better than random based approaches?},
  author={Bu{\ss}, Oliver and Rudat, Jens and Ochsenreither, Katrin},
  journal={Computational and structural biotechnology journal},
  volume={16},
  pages={25--33},
  year={2018},
  publisher={Elsevier}
}
\bibliographystyle{plain}

\section*{Data Availability}
Both the MolBench dataset (CSV format) and associated evaluation code can be accessed from Github (\url{https://github.com/InternScience/MolClaw}).

\section*{Code Availability}
The skills adopted in MolClaw, together with the corresponding configuration instructions, are also available at GitHub (\url{https://github.com/InternScience/MolClaw}).

\end{document}